\documentclass[sigconf]{aamas} 

\usepackage{balance} 

\usepackage{wrapfig}

\usepackage{enumitem}

\usepackage{algorithm}
\usepackage[noend]{algpseudocode}
\algnewcommand{\LeftComment}[1]{\Statex \(\triangleright\) #1}

\usepackage{multirow}
\usepackage{tabularx}
\usepackage{diagbox}
\usepackage{adjustbox}

\usepackage{pythonhighlight}

\usepackage{placeins}

\newcommand{\gD}{\mathcal{D}}
\newcommand{\I}{\mathbb{I}}
\newcommand{\Qtarget}{Q_{\hat{\theta}}}
\newcommand{\Q}{Q_{\theta}}
\newcommand{\V}{V_\psi}
\newcommand{\Lt}{\mathcal{L}_2^e}
\newcommand{\policy}{\pi_\phi}
\newcommand{\wiv}{w_V^i}
\newcommand{\bv}{b_V}
\newcommand{\Vi}{V_{\psi^i}}
\newcommand{\wiq}{w_Q^i}
\newcommand{\bq}{b_Q}
\newcommand{\Qi}{Q_{\theta^i}}
\newcommand{\wiqtarget}{\hat{w}_Q^i}
\newcommand{\bqtarget}{\hat{b}_Q}
\newcommand{\Qitarget}{Q_{\hat{\theta}^i}}

\newcommand{\eq}{Eq.~}

\DeclareMathAlphabet{\mathsfit}{\encodingdefault}{\sfdefault}{m}{sl}
\SetMathAlphabet{\mathsfit}{bold}{\encodingdefault}{\sfdefault}{bx}{n}

\def\gA{{\mathcal{A}}}

\def\gD{{\mathcal{D}}}

\def\gM{{\mathcal{M}}}
\def\gN{{\mathcal{N}}}

\def\gR{{\mathcal{R}}}
\def\gS{{\mathcal{S}}}

\newcommand{\E}{\mathbb{E}}

\newcommand{\R}{\mathbb{R}}

\newcommand{\e}[1]{\text{e}^{#1}}

\setcopyright{ifaamas}
\acmConference[AAMAS '24]{Proc.\@ of the 23rd International Conference
on Autonomous Agents and Multiagent Systems (AAMAS 2024)}{May 6 -- 10, 2024}
{Auckland, New Zealand}{N.~Alechina, V.~Dignum, M.~Dastani, J.S.~Sichman (eds.)}
\copyrightyear{2024}
\acmYear{2024}
\acmDOI{}
\acmPrice{}
\acmISBN{}

\acmSubmissionID{AAMAS2024 Submission 95}

\title[Offline MARL]{A Model-Based Solution to the Offline Multi-Agent Reinforcement Learning Coordination Problem}

\author{Paul Barde}
\affiliation{
  \institution{Mila - Quebec AI Institute \\ McGill University}
  \city{Montreal}
  \country{Canada}
  }
\email{bardepau@mila.quebec}

\author{Jakob Foerster}
\affiliation{
  \institution{University of Oxford}
  \city{Oxford}
  \country{United Kingdom}}

\author{Derek Nowrouzezahrai}
\affiliation{
  \institution{Mila - Quebec AI Institute \\ McGill University}
  \city{Montreal}
  \country{Canada}
  }

\author{Amy Zhang}
\affiliation{
  \institution{The University of Texas at Austin \\ Meta AI - FAIR}
  \city{Austin}
  \country{USA}
  }

\begin{abstract}
Training multiple agents to coordinate is an essential problem with applications in robotics, game theory, economics, and social sciences. However, most existing Multi-Agent Reinforcement Learning (MARL) methods are online and thus impractical for real-world applications in which collecting new interactions is costly or dangerous. While these algorithms should leverage offline data when available, doing so gives rise to what we call \textit{the offline coordination problem}.
Specifically, we identify and formalize the \emph{strategy agreement (SA)} and the \emph{strategy fine-tuning (SFT)} coordination challenges, two issues at which current offline MARL algorithms fail. Concretely, we reveal that the prevalent model-free methods are severely deficient and cannot handle coordination-intensive offline multi-agent tasks in either toy or MuJoCo domains. To address this setback, we emphasize the importance of inter-agent interactions and propose the very first model-based offline MARL method. Our resulting algorithm, Model-based Offline Multi-Agent Proximal Policy Optimization (MOMA-PPO) generates synthetic interaction data and enables agents to converge on a strategy while fine-tuning their policies accordingly. This simple model-based solution solves the coordination-intensive offline tasks, significantly outperforming the prevalent model-free methods even under severe partial observability and with learned world models. 
\end{abstract}


\newcommand{\BibTeX}{\rm B\kern-.05em{\sc i\kern-.025em b}\kern-.08em\TeX}

\makeatletter
\gdef\@copyrightpermission{
	\begin{minipage}{0.3\columnwidth}
		\href{https://creativecommons.org/licenses/by/4.0/}{\includegraphics[width=0.90\textwidth]{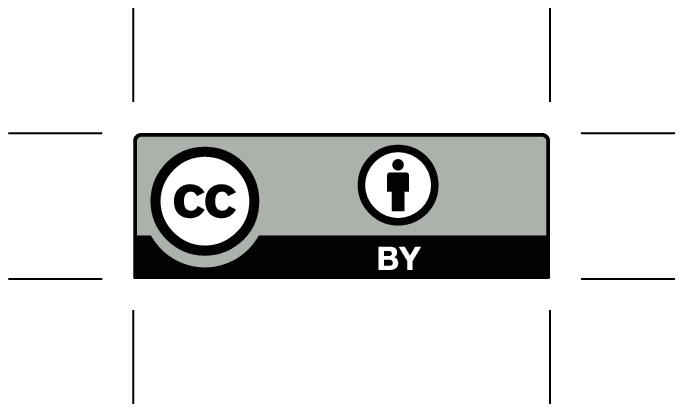}}
	\end{minipage}\hfill
	\begin{minipage}{0.7\columnwidth}
		\href{https://creativecommons.org/licenses/by/4.0/}{This work is licensed under a Creative Commons Attribution International 4.0 License.}
	\end{minipage}
	\vspace{5pt}
}
\makeatother

\begin{document}


\pagestyle{fancy}
\fancyhead{}


\maketitle 


\section{Introduction}
Multi-agent problems are ubiquitous in real-world scenarios, including traffic control, distributed energy management, multi-robot coordination, auctions, marketplaces, and social networks \cite{dresner2004multiagent, palanisamy2020multi, steeb1981distributed, boyan1993packet, varga1994integrating, huang1995agent, brauer1998multi, fischer1993sophisticated, cederman1997emergent, grand1997creatures,crawford1982strategic, panait2005cooperative}. This makes the development of efficient multi-agent algorithms a crucial research area in artificial intelligence and machine learning with substantial implications in various fields including robotics, game theory, economics, and social sciences \citep{vinitsky2022nocturne,shi2021offline,zhang2019cooperative, liu2021online,li2021scalable}. However, existing methods are mostly online and require interacting with the environment throughout learning which often makes them costly or even dangerous for real-world applications \cite{levine2020offline}. In contrast, offline Reinforcement Learning (offline RL) obviates the need for interactions with the environment as it allows learning from existing datasets that do not have to be collected by experts. It is therefore well suited to tasks where: a) we cannot afford to materialize the situation in practice, b) it is unfeasible to build a simulator, and c) there exist datasets of realizations of such situations.
Consequently, we hypothesize that offline multi-agent approaches will be key for tackling real-world multi-agent problems. Let us imagine for instance trying to understand how autonomous actors (i.e., governments, international organizations, industries, etc.) must maneuver to reduce the severity of a worldwide pandemic while preventing economic collapse. It goes without saying that: a) starting pandemics is not a viable way to gain real-world practice; b) building a simulator is a colossal task that would suggest emulating our society and its economy; and c) the impact of past decisions (such as implementing lockdown policies, travel restrictions, and vaccination campaigns) on the unfolding of the pandemics and the economy is well-documented. Hopefully, these records can be used to derive new strategies in the future.
\begin{figure*}[ht]
    \centering
    \begin{tabular}{ccc}
        \includegraphics[height=0.15\textheight]{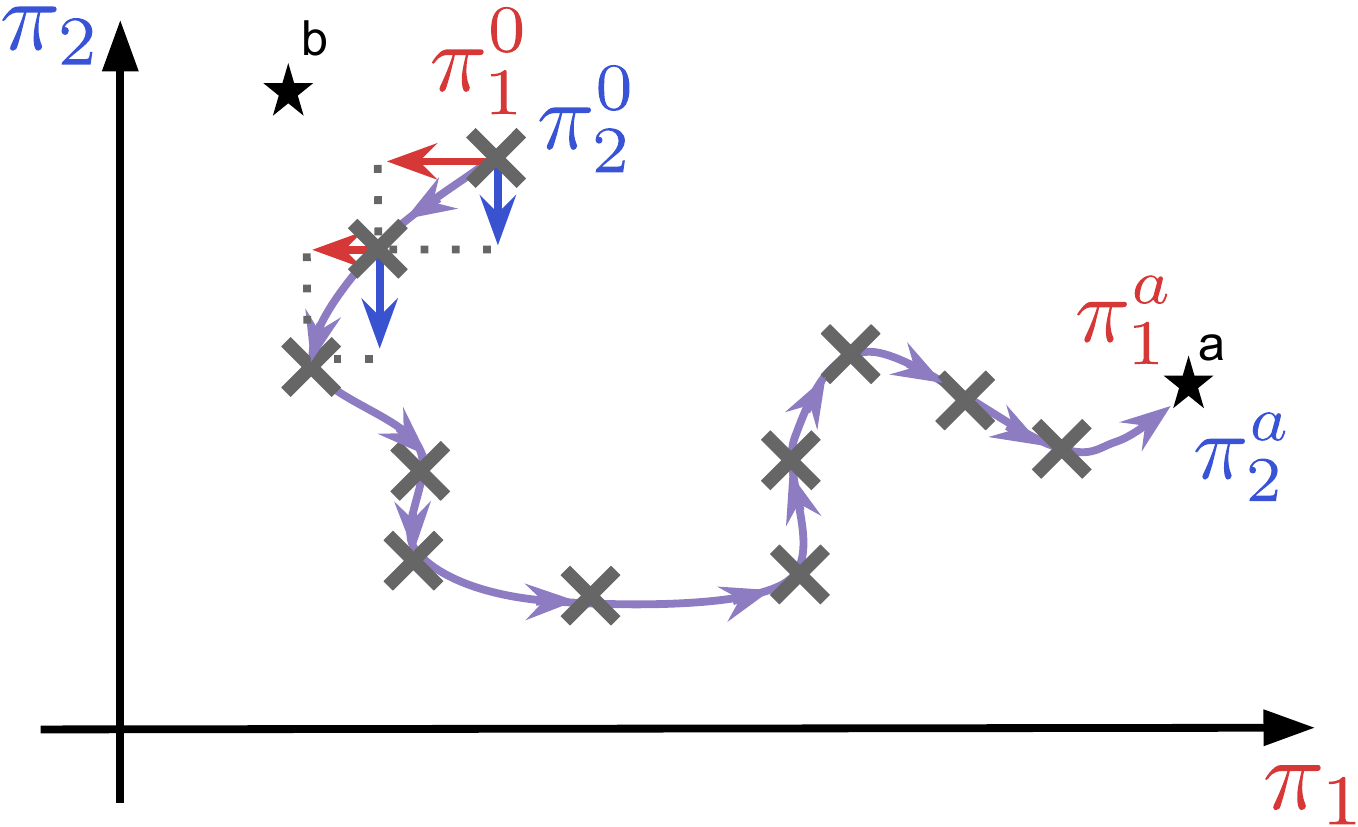} & 
        \includegraphics[height=0.15\textheight]{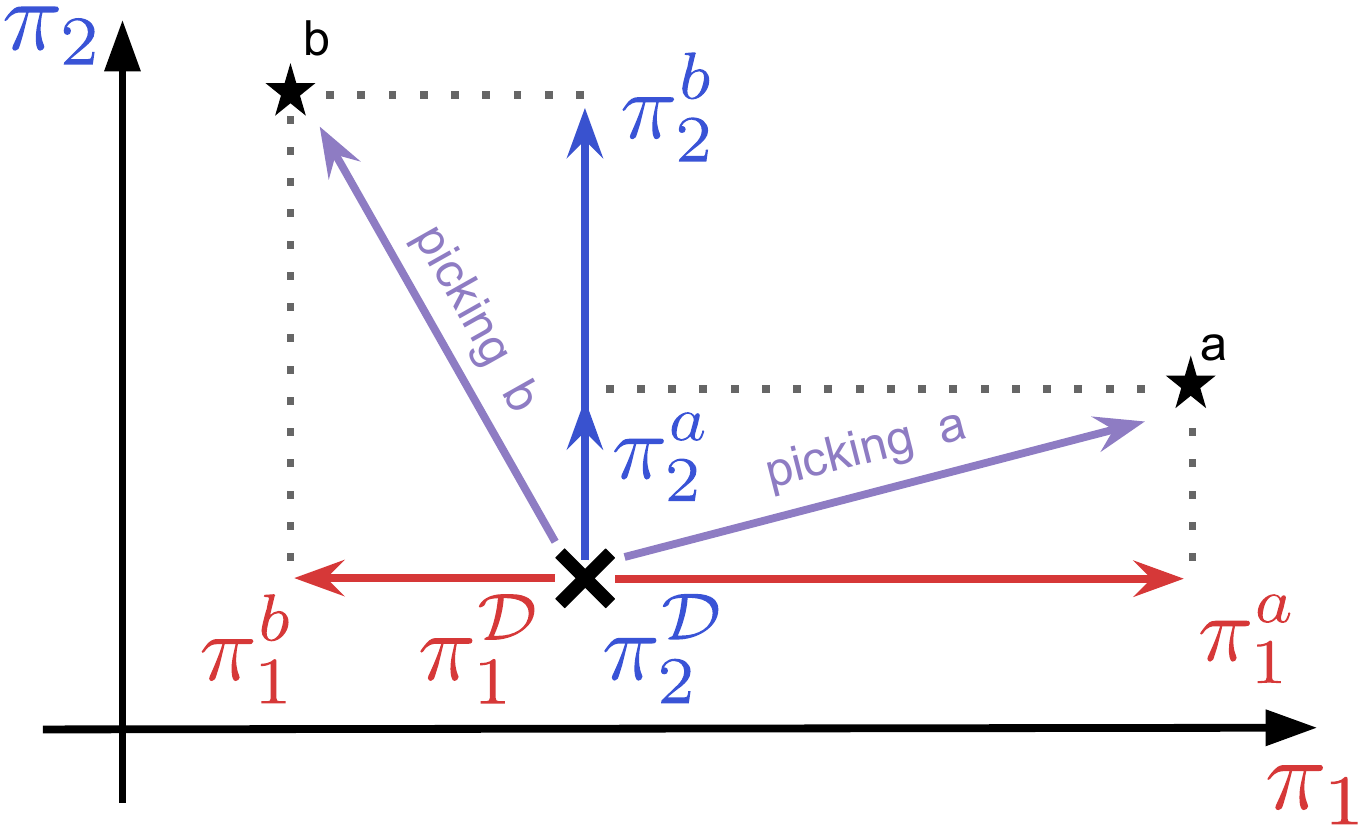} & \includegraphics[height=0.14\textheight]{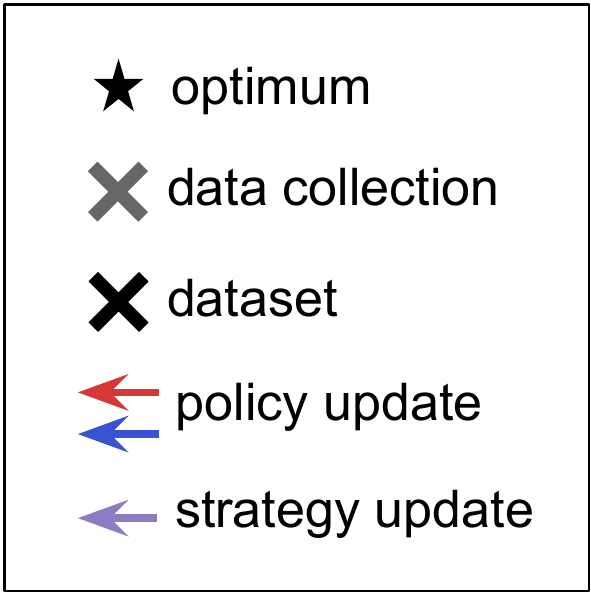}\\
         (a) online learning & (b) offline learning &
    \end{tabular}
    \caption{Comparing online learning and offline learning in policy space to illustrate the offline coordination problem.\\ (a) During online learning, agents continuously interact using their current policies and collect new data that informs the next update of the co-evolution from $(\pi_1^0, \pi_2^0$) to $(\pi_1^a, \pi_2^a$). (b) During offline learning, agents cannot collect new data and thus, they can only estimate updates from the dataset of interactions (here collected with $\pi_1^{\gD}$ and $\pi_2^{\gD}$). To reach an optimal strategy agents must (1) agree on which optimum to target between $\star^a$ or $\star^b$, -- i.e. solve \textit{strategy agreement} --, and (2) respectively derive the policy corresponding to that strategy ($\pi_1^{\star j}$ and $\pi_2^{\star j}$), -- i.e. solve \textit{strategy fine-tuning}.}
    \label{fig:ocp}
\end{figure*}

\paragraph{\textbf{The offline coordination challenge.}}
In general, actors such as individuals, organizations, robots, software processes, or cars are self-governed and ultimately act autonomously. However, during the offline learning phase, it is reasonable to assume that learners can share pieces of information, which makes the Centralized Training Decentralized Execution (CTDE) formulation a natural approach for the offline training regime. Unfortunately, as shown in this work, simply combining CTDE MARL and offline RL methods on a dataset of multi-agent interactions does not necessarily yield policies that perform well together. This is what we call \emph{the offline coordination problem}: 1) multi-agent solutions require agents to coordinate, that is to act coherently as a group such that individual behaviors combine into an efficient team strategy; 2) learning to coordinate is particularly challenging in the absence of interactions during training.

In this work, we propose that multi-agent coordination can be decomposed into two distinct challenges. First, since most multi-agent tasks allow multiple optimal team strategies \citep{boutilier1996planning} -- some of which may only vary on how they break symmetries present in the task \citep{hu2020other} -- agents must collectively select one strategy over another such that they individually converge toward coherent behaviors. We refer to this coordination challenge as \emph{Strategy Agreement (SA)}. Additionally, for a chosen team strategy, agents have to precisely calibrate and adjust their behaviors to one another in order to reach the corresponding optimal group behavior. We call this \emph{Strategy Fine-Tuning (SFT)}. During online learning, agents continuously interact together in the environment, therefore, changes due to one agent's local optimization directly impact other agents' experiences and teammates are able to adapt: coordination occurs through interactive trial and error. Conversely, when learning from a fixed dataset of interactions, agents cannot interact with other agents to generate new data that explicitly measures the outcomes of current policies in the environment (i.e., which global strategy is being chosen and how individual behaviors blend together). Therefore, it is difficult for offline learners to coordinate. Figure~\ref{fig:ocp} illustrates this in policy space: (a) During online learning, agents continuously interact to collect up-to-date data that inform the optimization of both individual policies and team strategy. This allows agents to converge to a global optimum by co-adapting and improving on each other's updates. (b) In the offline setting however, agents must independently decide towards which of the two optima they aim to converge (i.e., strategy agreement between $\pi_i^{\star a}$ or $\pi_i^{\star b}$, $i=1,2$). Assuming they pick $\star a$, they must then derive their corresponding optimal policy (i.e., strategy fine-tuning toward $\pi_i^{\star a}$, $i=1,2$). Crucially, offline agents must rely only on interactions present in the dataset (corresponding to policies $\pi_i^\gD$, $i=1,2$), thus likely without interactions corresponding to their past or current policies.  

Current methods \citep{yang2021belive, jiang2021offline, pan2022plan} deal with offline MARL by simply extending single-agent offline RL to the multi-agent setting. To do so, they either consider that agents are independent learners, or they leverage the CTDE paradigm. Provided that the dataset has enough coverage, it is in theory possible for CTDE agents to learn the different optimal value functions and policies. Yet, as we will see in Section~\ref{sec:results}, agents may still fail to agree on which team strategy to pick. We highlight such offline coordination failure in an offline version of the well-established Iterated Coordination Game \citep{boutilier1996planning}. The crux of \textit{offline} MARL is therefore that learners cannot interact together in the environment to collect new data that measures how individual policies blend together and what are the outcomes of the current team strategy. This relates to the absence of interactions during learning and is not a centralization issue.
Thus, we motivate the need for generating synthetic data that measure the outcomes of the team's current strategy to allow for coordinated agents. 

We propose MOMA-PPO, a simple model-based approach that generates synthetic interactions. We show that it allows for offline coordination even in complex Multi-Agent MuJoCo (MAMuJoCo) \cite{peng2021facmac} tasks with partial observability and learned world model. Across tasks and domains, MOMA-PPO significantly outperforms the offline MARL baselines. Our contributions are:
\vspace{-22pt}
\begin{itemize}[itemsep=4pt, left=6pt]
    \item identifying and defining the offline coordination problem, an issue that has been overlooked by the offline MARL community;
    \item proposing benchmarks in the form of new datasets and partially observable multi-agent tasks that test for offline coordination;
    \item showing that current methods, which are all model-free, fail at offline coordination even in simple environments;
    \item suggesting a link between the offline coordination problem and the lack of inter-agent interactions during learning that is inherent to model-free offline approaches;
    \item proposing to address the coordination problem with the first-ever model-based offline MARL method;
    \item experimentally validating the benefits of coordinating by interacting through a world model.
    \vspace{10.5pt}
\end{itemize}

\section{Background}
\label{sec:background}
We consider the formalism of Dec-POMDP \citep{nair2003taming} with states $s_t\in \gS$. There are $i=1,...,N_A$ agents that partially observe this state from their stochastic observation function $s^i_t\sim O^i(s^i|s_t)$ and choose action $a^i_t\in\gA^i$ at every time step. Each agent has an action-observation history given by $h_t^i=\{s_0^i, a_0^i, ...s_t^i\}$ and picks its action $a_t^i$ using its stochastic policy $\pi_i(a^i|h_t^i)$. The environment's transition depends on the joint action $a_t$ such that $s_{t+1} \sim P(s'|s_t, a_t)$. The game is fully cooperative so all agents receive the same reward and $r_t^i = r_t = r(s_t,a_t) \in \R$ $\forall i$. The agent's objective is therefore to maximize the expected team return $J=\E_\tau R(\tau)$ where $R(\tau)=\sum_t\gamma^tr_t$ with discount factor $\gamma$ and $\tau=\{s_0, a_0, r_0, ...s_F\}$ is a trajectory with absorbing state $s_F$. Note that some trajectories can become arbitrarily long in which case we truncate them and use the value of the last state as an estimation of the return-to-go.

We adopt the CTDE framework \cite{lowe2017multi, foerster2018counterfactual}, where at training time, individual observations, policies, and value functions are available to all the agents. For simplicity, we assume access to the global state $s_t$ and the observation functions $O^i$. Yet, this is not a requirement and it is always possible to define the global state as the concatenation of the agents' observations, i.e., $s_t = \{s_t^1, ..., s_t^{N_A}\}$. In most problems, which are special cases of Dec-POMDPs referred to as Dec-MDPs, such concatenation will fully observe the environment. At execution, agents must act independently and the joint policy is approximated by individual decentralized policies as
$
a_t \sim \pi(a|s_t) \approx \prod_{i=0}^{N_A} \pi_i(a^i|h^i_t)
$.

Finally, we consider Offline Learning \cite{fujimoto2019off, levine2020offline}, where agents have only access to a fixed dataset of trajectories $\gD$ and cannot collect additional interactions with the environment during learning.


\section{Method}
\label{sec:method}
In this work, we propose MOMA-PPO, a Dyna-like \citep{sutton1990integrated} model-based approach to multi-agent CTDE offline learning that relies on PPO \cite{schulman2017proximal}. The method can be decomposed into two steps: 1) learning a world model from the dataset, and 2) using the world model to train the agents' policies. 

\subsection{Learning a centralized world model ensemble}

MOMA-PPO leverages the CTDE assumptions to learn centralized models that predict the next state, reward, and termination condition from the current state and actions. When learning in an approximate world model, RL agents might learn to exploit the world model's reconstruction inaccuracies to reap more rewards in simulation, eventually producing incoherent behaviors that perform poorly in the real world \cite{ha2018world}. One way to avoid this is to penalize the agents for going into regions of the state-action space where the world model is uncertain about its predictions \cite{yu2020mopo}. Learning an ensemble of models enables estimating the world model's epistemic uncertainty due to the limited amount of learning data in the offline dataset. Each model comprises two diagonal Gaussians $\gN(\mu_T, \sigma_T^2)$ and $\gN(\mu_r, \sigma_r^2)$ that respectively model the next state $s'$ and the reward $r$. Models also predict whether or not the next state is terminal using a Bernoulli distribution Bern($p_d$). Distributions' $\mu_T,  \mu_R, \sigma_T$, $\sigma_R$, and $p_d$ are parametrized by neural networks conditioned on the current global state $s$ and the joint action $a$. The parameters are learned from the offline dataset $\gD$ using Gaussian negative log-likelihood for $\gN(\mu_T, \sigma_T^2)$ and $\gN(\mu_r, \sigma_r^2)$, and binary cross-entropy for Bern($p_d$).
In practice, we train $N_m = 7$ models and keep the best $N=5$ based on their average validation accuracy across the next states and rewards \cite{yu2020mopo}. 
We estimate the epistemic uncertainty of the reward using the variance of the predicted rewards across the ensemble: 
$$
\epsilon_r = \frac{\sum_{m=1}^{N}(\hat{r}_m-\bar{r})^2}{N-1}, \qquad \bar{r} = \frac{\sum_{m=1}^{N} \hat{r}_m}{N}.
$$

We also estimate the epistemic uncertainty of the general prediction by concatenating the next state and the reward and computing the Frobenius norm of the ensemble covariance matrix:
\begin{align*}
& \epsilon_g = ||\text{cov}(x_i, x_j)||_{F} \quad, \\
& \text{cov}(x_i, x_j) = \frac{\sum_{m=1}^{N}(\hat{x}_{i,m}- \bar{x_j})(\hat{x}_{j,m}-\bar{x_i})}{N-1},    
\end{align*}
where $x_i$ and $x_j$ are components of the vector resulting from the concatenation of the predicted next state vector $\hat{s}'$ and the predicted reward scalar $\hat{r}$. At this point, we define a world model based on the ensemble such that: 
\begin{equation*}
    \hat{s}_{t+1}, \bar{r}_t, \bar{f}_t, \epsilon_{r,t}, \epsilon_{g,t} \sim \gM(\cdot| s_t, a_t), 
\end{equation*}
where $\bar{f}_t$ is a mask equal to 0, if the model predicts that we reached an absorbing state, and 1 otherwise. $\bar{r}_{t}$ is the mean predicted reward across the ensemble and $\bar{f}_t$ results from a majority vote between the members of the ensemble. The mean state would likely be out-of-distribution and lack the structure of real states which would impede learning and evaluation. Therefore, $\hat{s}_{t+1}$ is instead sampled uniformly amongst each ensemble member's next state. Finally, to avoid unrealistic values, $\bar{r}_{t}$ and $\hat{s}_{t+1}$ are clipped to the minimum bounding box of the offline dataset while uncertainties estimations are limited to a specified threshold $l_\epsilon$: 
\begin{align*}
&\begin{aligned}
    &\min_{r\in\gD} r \leq \bar{r}_t \leq \max_{r\in\gD} r\\
    &\min_{s_i\in\gD} s_i \leq \hat{s}_{t+1, i} \leq \max_{s_i\in\gD} s_i
\end{aligned} \quad \forall i \in [0, q] \, | \, \gS \subset \R^q,\\ 
    &\text{and}\quad \epsilon_r \leq l_\epsilon, \, \epsilon_g \leq l_\epsilon.
\end{align*}

\subsection{Model-based Offline Multi-Agent PPO (MOMA-PPO)}
\setlength{\columnsep}{12pt}
\begin{wrapfigure}{r}{0.2\textwidth}
\vspace{-16pt}
    \centering
    \includegraphics[width=0.18\textwidth]{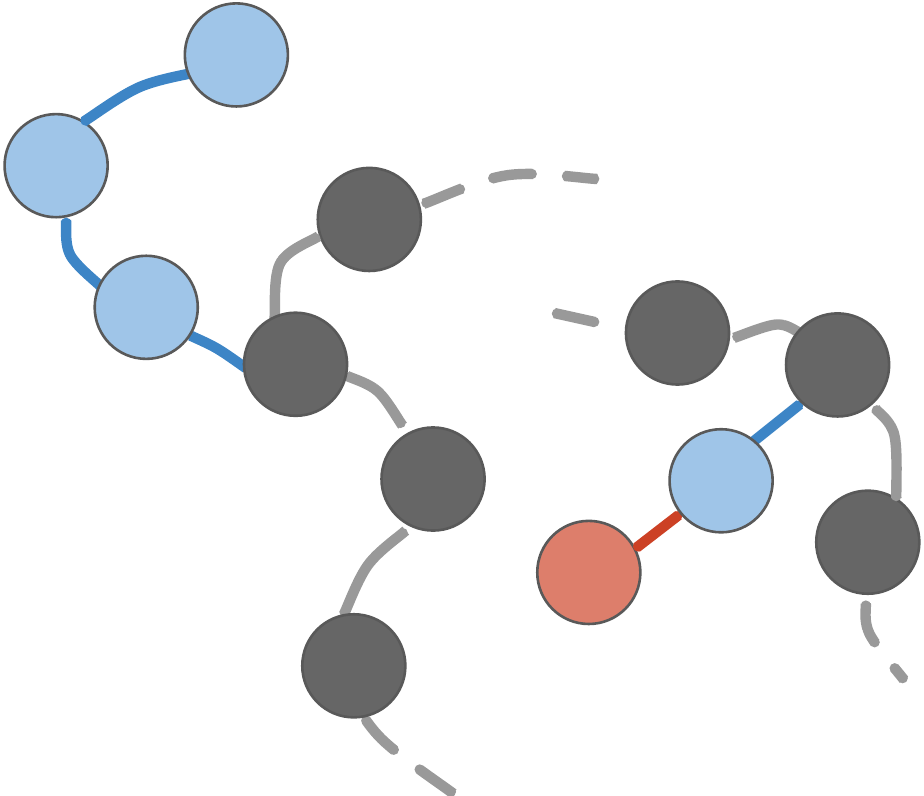}
    \caption{\textbf{Model-based rollouts generation} (blue) from dataset's states (grey). Red denotes early termination and $k=3$.}
    \label{fig:rollouts}
    \vspace{-6pt}
\end{wrapfigure}

Once $\gM$ has been trained on the offline dataset $\gD$, it can be used to train online reinforcement learning algorithms in a Dyna-like manner. Here, we use MAPPO, a CTDE multi-agent version of PPO \cite{yu2022surprising}.

The synthetic data used to train the PPO policies is collected by sampling states from the offline dataset $\gD$ and using the current policies $\pi_i$ alongside the world model $\gM$ to generate PPO's training rollouts of size $k$. Terminating a rollout when the world model uncertainty exceeds $l_\epsilon$ allows for adaptative rollout length and avoids training the policies on unfeasible data. Rollouts are always terminated with a timeout mask $\zeta_t$ such that: 
$$
\zeta_{t} = 1 - \I(t=k-1 \cup \epsilon_{g,t} \geq l_\epsilon),
$$
where $\I$ is the indicator function. Note that the timeout mask $\zeta_{t}$ is different from the model-predicted mask $f_t$ which is solely related to the environment and indicates terminal absorbing states that occur for instance when the agent dies or reaches the goal. On the PPO side, we adapt the generalized advantage estimation \cite{schulman2015high} to account for the timeouts $\zeta_{t}$ and ensure that there is no accumulation across rollouts while computing returns. We use the value of the last state as an estimation of the return-to-go (see Appendix \ref{app:method}). 

The generation of length $k=3$ rollouts is illustrated in Figure~\ref{fig:rollouts} where it can be seen that a rollout is interrupted early because the world model was unreliable when generating the associated state (shown in red). 

In addition to the adaptive rollout length, the model epistemic uncertainty is used to penalize the agents so they avoid exploiting the world model in poorly reconstructed regions of the state space. The final uncertainty-penalized reward is given by:
$$
\tilde{r}_t = \bar{r}_t - \lambda_r \epsilon_r - \lambda_g \epsilon_g,
$$
where $\epsilon_r$ and $\epsilon_g$ are hyperparameters that weigh the severity of the penalty.

\begin{figure}[H]
\vspace{-16pt}
\begin{algorithm}[H]
\caption{MOMA-PPO}
\small
\label{alg:practical}
\begin{algorithmic}[0]
\Require offline dataset $\gD$, world model $\gM$, rollout horizon $k$, rollout batch size $b$, uncertainty penalty coefficients $\lambda_r$ and $\lambda_g$, uncertainty threshold $l_\epsilon$,  MAPPO agents.
  \State Initialize MAPPO policies $\pi_i$ and value function $V$.
    \For{epoch $1, 2, \dots$}
	  \LeftComment{Generate synthetic data}
            \State Initialize an empty rollout buffer $\gR \leftarrow \varnothing$. 
		  \For{$1, 2, \dots, b$ (in parallel)}
		        \State Sample history $h_t = \{h^i_t\}_{i=1}^{N_A}$ from $\gD$.
		        \For{$j = t, t+1, \dots, t + k-1$}
		            \State Sample $a^i_{j} \sim \pi_i(a^i|h^i_{j}) \, \forall i$.
		            \State Sample $\hat{s}_{j+1}, \bar{r}_{j}, \bar{f}_{j}, \epsilon_{r, j}, \epsilon_{g, j} \sim \gM(\cdot|s_{j},a_{j})$.
		            \State Compute $\tilde{r}_j = \bar{r}_j - \lambda_r \epsilon_r - \lambda_g \epsilon_g$.
                        \State Compute $\zeta_{j} = 1 - \I(j=t+k-1 \cup \epsilon_{g,j} \geq l_\epsilon)$
		            \State Add sample $(h_j, a_j, \tilde{r}_j, \hat{f}_j, \zeta_j, \hat{s}_{j+1})$ to $\gR$.
                        \State Get $h^i_{j+1}$ from $h^i_{j}, \, a^i_{j}$ and $\hat{s}_{j+1}$. Set $s_{j+1} = \hat{s}_{j+1}$.
                        
                        \label{line:2}
		        \EndFor
		    \EndFor
        \LeftComment{Train agents}
		    \State Use synthetics rollouts in $\gR$ to train policies $\pi_i$ and value 
                \State function $V$ with MAPPO.
		\EndFor
 \\
\Return multi-agent policies $\pi_i$.
	\end{algorithmic}
\end{algorithm}
\vspace{-18pt}
\end{figure}

\paragraph{\textbf{Practical considerations and limitations.}}
As stated in Section~\ref{sec:background}, we assume CTDE and that the global state $s_t$ fully observes the environment, therefore we do not equip the world model with memory. The agents, on the other hand, only have access to partial observations and must rely on action-observation histories $h^i_t$. In practice, we restrict action-observation histories to 10 steps in the past (either from the dataset or from the generated rollouts) and process them with one layer of self-attention \cite{vaswani2017attention} followed by one layer of soft-attention \cite{bahdanau2015neural}. The resulting embeddings are concatenated to the agent's current state, and for simplicity, we abuse notation by denoting this ``memory enhanced" state with $h^i_t$. 

In MAPPO our centralized value function uses the QMIX value-decomposition \citep{rashid2018qmix} with $w^i(s_t)$ and $b(s_t)$ provided by a learnable neural network: \mbox{$V(s_t)\triangleq \sum_i w^i(s_t)V^i(h^i_t) + b(s_t)$}.

Finally, it is important to note that the task of the MOMA-PPO agents is quite different from the task of the agents that generated the dataset. First, the MOMA-PPO agents' initial state distribution is now the dataset's state distribution. Then the reward of the task has been altered to account for the model uncertainty. Last but not least, agents are only allowed to stray $k$ steps away from the dataset's coverage. While this restriction mitigates world model abuse, it can also prevent the agents from discovering goals that are further away from the offline data. Our resulting model-based offline multi-agent method is illustrated in Algorithm~\ref{alg:practical} and more details are provided in Appendix~\ref{app:method}.

\section{Related Work}
\paragraph{\textbf{Multi-agent coordination.}}
Coordination has been a challenge of interest since the early works on cooperative MARL \citep{boutilier1996planning, claus1998dynamics, littman2001friend, brafman2002efficient, chalkiadakis2003coordination} and has consistently been a central focus of the multi-agent literature \citep{zhang2013coordinating, lowe2017multi, jaques2018intrinsic, lerer2019learning}. While different works consider different aspects of coordination -- such as behavior predictability and synchronous sub-policy selection \citep{roy2020promoting}, structured team exploration \citep{mahajan2019maven} or the emergence of communication and cooperative guiding \cite{lazaridou2017multiagent, woodward2020learning, barde2022learning} -- our definition of coordination is closest to the seminal work of \cite{boutilier1996planning}. Indeed, we consider coordination in terms of agents agreeing to individually follow the same team strategy (that is a policy over joint actions) and finetuning their behavior to one another in tasks where multiple distinct optimal team strategies exist. 
A similar notion of coordination has been used in the \textit{zero-shot coordination problem} investigated by \cite{hu2020other} where agents are trained so that they are able to perform with agents they have never seen before. Yet, while their focus is on deriving standardized coordinated strategies that can generalize to unseen teammates, coordination is still learned through online interactions.

Coordination with teammates without direct interactions is often referred to as \emph{ad-hoc coordination} \cite{stone2010ad, barrett2011empirical}. 
Recent works assume access to offline demonstrations of the teammates’ behaviors. These can be used to guide the agent’s self-play training toward adopting the appropriate equilibrium (or “social conventions”) of its future teammates \cite{lerer2019learning, tucker2020adversarially}. Similarly, \cite{carroll2019utility} uses offline
data to learn a model of the teammate’s behavior and use it to train the agent to coordinate with that ally. In ad-hoc coordination, teammates' behaviors are fixed and can be estimated a priori from the dataset: the learner merely has to identify the team strategy and adopt it. Conversely, in our offline coordination setting, all the agents are learning and therefore have unknown changing behaviors: they must identify the different potential team strategies and agree on which one to follow (\emph{strategy agreement}). Simultaneously, they must finetune their behaviors to one another in order to reach this team policy (\emph{strategy fine-tuning}). All this without being able to interact with other agents or the environment. 

\paragraph{\textbf{Offline MARL}}
Recent works have been investigating offline solutions to the MARL problem. All of these methods build on model-free single-agent approaches and constrain the policy to stay in the dataset's distribution by using either SARSA-like schemes (such as ICQ \cite{yang2021belive} and IQL \citep{kostrikov2021offline}) or policy regularization (such as CQL \cite{kumar2020conservative} and TD3+BC \cite{fujimoto2021minimalist}). 

Some methods investigate specific modifications to improve performance in the multi-agent setting. For instance, in the decentralized setting, MABCQ \cite{jiang2021offline} enforces an optimistic importance sampling modification that assumes that independent agents will strive toward similar high-rewarding states, yet since this does not discriminate between which high-rewarding state to favor, the strategy agreement issue remains. For discrete actions spaces problems, \cite{tseng2022offline} propose a Transformer-based approach that learns a centralized teacher and distills its policy into independent student policies.
Finally, OMAR \cite{pan2022plan} proposes to alleviate miscoordination failure in offline MARL by adding a zeroth order optimization method on top of multi-agent CQL, achieving state-of-the-art performance on a variety of tasks. We share these works' goal of learning coordinated and efficient multi-agent teams in the offline setting. Yet, we believe that interacting learners and agents are essential to coordination and therefore take a different approach by focusing on model-based methods rather than model-free ones.  

\paragraph{\textbf{Offline model-based RL}}
Model-based approaches have been investigated in the single-agent offline RL setting. Notoriously, MOPO \citep{yu2020mopo} proposed to learn an ensemble-based world model and use it to generate rollouts from the offline dataset to train a SAC agent \cite{haarnoja2018soft}. They also proposed an uncertainty-based reward penalty to prevent the learner from exploiting the model. MOReL \citep{kidambi2020morel} takes a similar approach but prevents model abuse by learning a pessimistic MDP in which states that are outside of the dataset coverage become absorbing terminal states. COMBO \citep{yu2021combo} proposed a similar but more conservative version of MOPO by using CQL instead of SAC and learning on both generated and dataset's states. Finally, ROMI \cite{wang2021offline} also uses model-free offline RL to derive a policy from a model-based augmented offline dataset, yet they enforce additional conservatism by learning a reverse policy and dynamics model to generate rollouts that lead to target states contained in the dataset. This mitigates against generating rollouts outside of the dataset's coverage.

We believe that offline RL algorithms are ill-suited to learn on non-stationary data such as the one generated by updating policies be it in a world model or in a real environment. Therefore, our method MOMA-PPO uses an online RL algorithm instead of an offline one. Additionally, to enforce conservatism and avoid world model exploitation we use both uncertainty penalty and early rollout termination. In that sense, MOMA-PPO unifies what would be multi-agent extensions of MOPO and MOReL. Yet, instead of penalizing aleatoric uncertainty as in MOPO, we focus on epistemic uncertainty and estimate it by monitoring the coherence between the different models in the ensemble. Section~\ref{sec:ablations} reports the benefits of using the epistemic uncertainty (due to the finite amount of training data) over the aleatoric uncertainty (from the environment's stochasticity). Also, MOMA-PPO's early rollout termination is done with timeouts rather than with MOReL's terminal states meaning that it does not penalize agents while still avoiding using unfeasible rollouts to train them. Finally, in contrast with MOReL and MOPO which use off-policy algorithms, MOMA-PPO is based on MAPPO, an on-policy method that has reliably achieved state-of-the-art performance in MARL tasks \cite{yu2022surprising}. It is therefore well suited to the multi-agent setting in which slight changes in a teammate's policy can drastically impact the overall group behavior and quickly make previous interactions obsolete. 

\paragraph{\textbf{Model-based multi-agent RL}}
In the online setting, model-based approaches aim to improve sample efficiency by reducing the number of interactions with the environment. Therefore, these methods use off-policy schemes and focus on how and when to collect additional environment data for refining the world model and the policies \cite{willemsen2021mambpo,zhang2021model,zhang2022marco}. In offline MARL, sample efficiency is not a consideration since the environment data has already been collected offline and additional samples are not an option. Yet, as we show here, model-based approaches can benefit multi-agent coordination in the offline setting by allowing multiple learners to interact through the world model.  

\section{Baselines, Environments, Tasks, and Datasets}
\begin{figure}[H]
    \centering
    \begin{tabular}{ccc}
        \resizebox{0.17\textwidth}{!}{
        \adjustbox{valign=c}{\begin{tabular}{|c|c|cc|}
        \cline{3-4}
        \multicolumn{2}{c|}{\multirow{2}{*}{}}  &  \multicolumn{2}{c|}{$a^2$ }\\
        \cline{3-4}
           \multicolumn{2}{c|}{}  &  $\leftarrow$ & $\rightarrow$ \\
            \hline
            \multirow{2}{*}{$a^1$}& $\leftarrow$ & 1,1 & 0,0 \\ 
             & $\rightarrow$ & 0,0 & 1,1 \\
            \hline
        \end{tabular}}}
         & 
         
         \adjustbox{valign=c}{\includegraphics[width=0.125\textwidth]{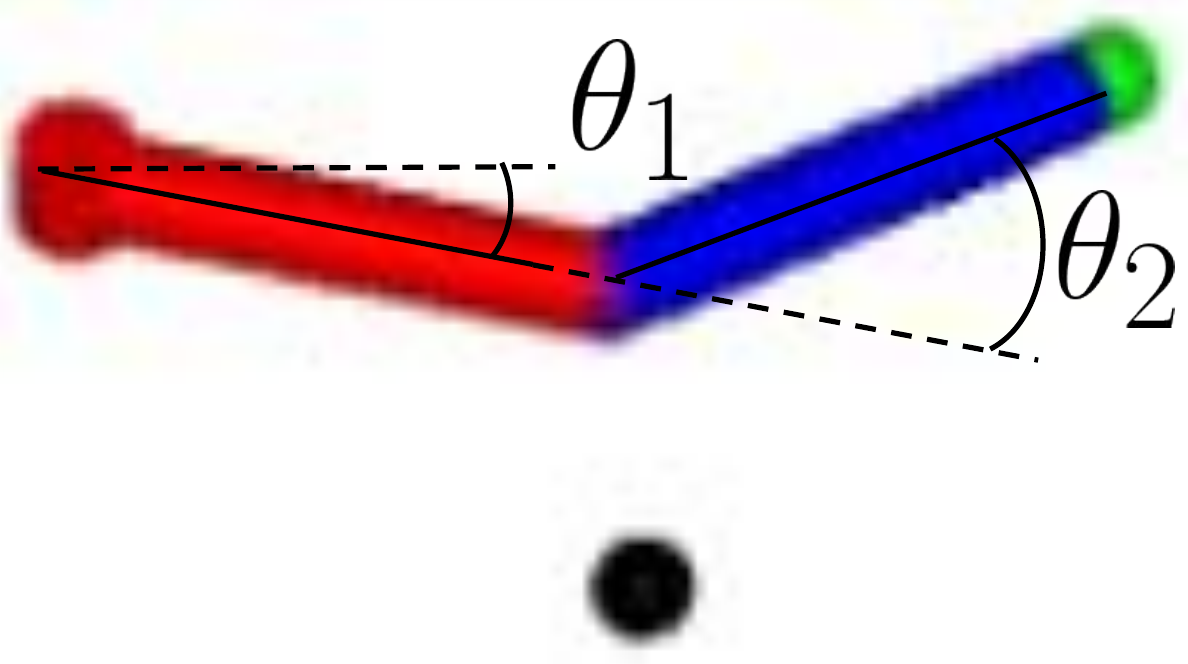}} & 

         \adjustbox{valign=c}{\includegraphics[width=0.125\textwidth]{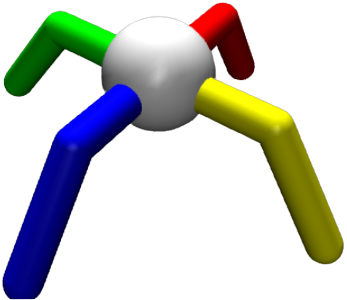}} \\
         (a) & (b) & (c)
    \end{tabular}
    \caption{\textbf{Environments illustrations.} (a) Pay off matrix of the Iterated Coordination Game. (b) Two-agent Reacher, red and blue agents respectively control the torque on $\theta_1$ and $\theta_2$. (c) Four-agent Ant, each agent controls a different limb (shown with different colors). In PO tasks, agents only observe the limb they control while the torso observations -- in white -- are available only to the yellow agent.}
    \label{fig:envs}
\end{figure}
\subsection{Baselines}
We compare with a large and varied array of baselines. First, we consider a centralized training \textit{and centralized execution} approach by observing the global state \textit{and selecting the joint action}. We use IQL \cite{kostrikov2021offline}, a state-of-the-art single-agent model-free offline RL algorithm for this setting. Considering centralized execution gives an upper bound on what can be achieved in terms of strategy agreement. Indeed, a single learner controls all the agents and can thus choose for the whole team the strategy to adopt.

Then, we extend IQL \cite{kostrikov2021offline} to the multi-agent setting by using the QMIX value decomposition on the $Q$ and $V$ value networks. This gives MAIQL, a very competitive model-free CTDE offline MARL algorithm that should allow for fine-tuning with additional online data after training. We refer to the finetuned version as MAIQL-ft and follow IQL's finetuning procedure \cite{kostrikov2021offline}: MAIQL-ft is first trained to convergence on the offline data and then finetuned for the same number of training steps by progressively introducing additional interaction data. At the end of finetuning, the replay buffer contains as many offline interactions as online ones. For completeness, we also consider MATD3+BC the CTDE multi-agent version of \cite{fujimoto2021minimalist} with QMIX.

Recent literature in model-free MARL \cite{de2020independent, lyu2021contrasting} and notably in the offline setting \cite{pan2022plan}, advocates for decentralized value functions and independent learners. Therefore, we consider several independent learner approaches, starting with Independent Behavioral Cloning (IBC). Despite its simplicity, BC \cite{pomerleau1991efficient} produces surprisingly efficient baselines for Imitation Learning and offline RL \cite{barde2020adversarial, spencer2021feedback}. Finally, we consider the independent learners extensions to \cite{kumar2020conservative} and \cite{fujimoto2021minimalist}, respectively ICQL and ITD3+BC, as well as the state-of-the-art model-free offline MARL method, OMAR \cite{pan2022plan}. 

\subsection{Offline Iterated Coordination Game}
To illustrate the strategy agreement coordination challenge, we propose an offline version of the Iterated Coordination Game \citep{boutilier1996planning} presented in Figure~\ref{fig:envs}~(a). Agents must pick the same direction in order to succeed and we investigate three offline datasets of interactions (see Table~\ref{tab:coord_game_dataset}). In the most favorable dataset, data is collected by coordinated agents that select the same option of going right most of the time. 
In the less favorable setting, agent 1 goes left most of the time while agent 2 is more likely to go right. In the neutral setting, agents act uniformly. However, each dataset does contain both coordinated and uncoordinated behaviors in which agents simultaneously choose the same -- respectively, different -- directions. 

It is therefore straightforward for a centralized critic to learn that $Q(\rightarrow, \rightarrow) = Q(\leftarrow, \leftarrow) = 1$ while $Q(\rightarrow, \leftarrow) = Q(\leftarrow, \rightarrow) = 0$, regardless of the specific dataset. Yet, decentralized actors remain unaware of whether they should go left or right since both strategies are equivalent and actors have no way of consistently picking one over the other.  
\begin{table}[H]
\vspace{-6pt}
    \centering
    \caption{Policies used to collect the datasets in the Iterated Coordination Game and resulting average scores of the datasets}
    \vspace{-6pt}
    \begin{tabular}{|c|c|c|c|}
    \cline{2-4}
    \multicolumn{1}{c|}{}& $P(a^1=\rightarrow)$ & $P(a^2=\rightarrow)$ & Avg. Score\\
    \hline
    favorable & 0.75 & 0.75 & 0.623 \\
    neutral & 0.5 & 0.5 & 0.502 \\
    unfavorable & 0.25 & 0.75 & 0.375\\
    \hline
    \end{tabular}
    \label{tab:coord_game_dataset}
    \vspace{-6pt}
\end{table}

\subsection{Offline MAMuJoCo}
Building upon D4RL \cite{fu2020d4rl} and MAMuJoCo \cite{peng2021facmac}, we propose offline multi-agent continuous control tasks with various datasets and full or partial observability. 

\paragraph{\textbf{Two-agent Reacher with a mixture-of-expert dataset.}}
To investigate strategy agreement in a more complex continuous control setting, we propose a two-agent version of the Reacher environment as shown in Figure~\ref{fig:envs}~(b). 
The offline dataset is collected as follows: in the first stage, we train online MAPPO on the fully observable two-agent Reacher task (every agent observes all the joint angles and velocities as well as the target position -- in black -- and the target to fingertip -- in green -- vector). Depending on the seed of the run, teams converge to counter-clockwise ($\theta_2\geq0$ as in Figure~\ref{fig:envs}~(b)) or clockwise ($\theta_2 \leq 0$) arm bends. Thus we can build a mixture-of-expert dataset by combining equal proportions of demonstrations from clockwise and counter-clockwise teams. Finally, we explore the impact of Full Observability (FO) versus Partial Observability (PO) by considering three types of observation functions: \textit{all-observant} (FO: every agent fully observes the environment), \textit{independent} (PO: each agent only sees the target and the velocity and angle of the joint it controls), and \textit{leader-only} (PO: both agents observes the two joints but only the red agent observes the target's position). Note that with PO no agent observes the target to fingertip vector. Thus, in \textit{leader-only}, only the leader can estimate whether or not the fingertip matches the target. In \textit{independent}, no agent has enough information to estimate this, yet both observe the target position. These tasks are very challenging and require agents to agree on following a specific convention (either clockwise or counter-clockwise arm bend) to reach a given target location and get rewards.

\paragraph{\textbf{Four-agent Ant.}} 
Similarly, we use a MAMuJoCo-like decomposition \cite{peng2021facmac} of the D4RL \cite{fu2020d4rl} offline ant task to make it multi-agent: each individual limb (composed of two joints) is controlled by a different agent. For the offline datasets, we use the single-agent D4RL datasets and consider two types of observation functions. For fully observable tasks, every agent observes the whole robot: torso observations (i.e., vertical position, orientation, angular and translational velocities) and the observations of all the limbs (i.e., angle and angular velocity of each joint). For the partially observable tasks, agents only observe the limb they control, and the torso observations are only made available to the yellow limb agent. PO tasks are very challenging in this case because only the yellow agent knows if the ant is moving in the correct direction and it must therefore learn to ``steer" the whole robot.

Table~\ref{tab:dataset_metrics} details all the datasets' score distributions as well are the reference agents' returns -- expert and random -- that are used to normalize performance. Note that ant datasets are from D4RL \cite{fu2020d4rl} (single agent datasets that we split into multi-agent observations and actions) while we generated the two-agent reacher mixture-of-expert dataset using MAPPO.
\begin{table}[H]
\vspace{-6pt}
    \centering
    \caption{Normalized measures of datasets' scores distributions and normalization performances.}
    \vspace{-6pt}
    \resizebox{0.47\textwidth}{!}{
    \begin{tabular}{|c|c||c|c|c|c||c|c|}
    \hline
    \multicolumn{2}{|c||}{\multirow{2}{*}{\diagbox{Datasets}{Scores}}} & min & mean & median & max &  expert & random \\ 
    \multicolumn{2}{|c||}{} & ($\%$) & ($\%$) & ($\%$) & ($\%$) &  return & return\\
    \hline
    reacher & expert-mix & 38.8 & 100.0 & 98.9 & 152.3 & -4.237 & -11.145 \\
    \hline
    \multirow{4}{*}{ant} & random & -3	& 6.4 & 7.2 & 10.3 & \multirow{4}{*}{3879.7} & \multirow{4}{*}{-325.6} \\
    & medium & -4.8 & 80.2 & 95.1 & 107.2 & & \\
    & full-replay & -22.4 & 72.0 & 77.8 & 134.3 & & \\
    & expert & -32.8 & 117.4 & 129.4 & 142.5 & & \\
    \hline
    \end{tabular}
    }
    \label{tab:dataset_metrics}
\end{table}


\section{Results}
\label{sec:results}
\begin{table}[h]
    \centering
    \caption{Teams' performances on the Iterated Coordination Game. MOMA-PPO is the only decentralized execution method to solve it for all the datasets.}
    \label{tab:toy-results}
    \vspace{-6pt}
    \resizebox{.47\textwidth}{!}{
    \begin{tabular}{|c|c||c||c||c|}
    \cline{2-5}
     \multicolumn{1}{c|}{}& IQL & MAIQL & IBC &  MOMA-PPO\\
     \cline{1-5}
     fav. & \textbf{1. $\pm$  0.} & \textbf{1. $\pm$ 0.} & \textbf{1. $\pm$ 0.} & \textbf{1. $\pm$ 0.} \\
     neutral & \textbf{1. $\pm$ 0.} & \textbf{0.9 $\pm$ 0.1} & 0.55 $\pm$ 0.11 & \textbf{1. $\pm$ 0.} \\
     unfav.  & \textbf{1. $\pm$ 0.} & 0. $\pm$ 0. & 0. $\pm$ 0. & \textbf{1. $\pm$ 0.}\\
     \hline
    \end{tabular}}
\end{table}

\begin{table*}[h]
    \centering
    \caption{Teams' performances on two-agent Reacher with mixture-of-experts dataset for different observation functions. Scores are normalized with expert and random performances. (a) Independent learners fail on datasets that contain a mixture of incompatible experts while MOMA-PPO (and to some extent MAIQL) are able to coordinate agents. (b) Current model-free methods are unable to adapt agents' behaviors while MOMA-PPO significantly outperforms the baseline across all settings.}
    \label{tab:reacher}
    \vspace{-6pt}
    \begin{tabular}{c}
    (a) Two-agent Reacher, mixture-of-experts dataset and full/partial observability \\
    
    \resizebox{.98\textwidth}{!}{\begin{tabular}{|c|c||c||c|c||c|c|c|c||c|}
    \hline
    \multicolumn{2}{|c||}{\multirow{3}{*}{\diagbox{Tasks}{Algorithms}}} & \multicolumn{7}{c||}{model-free} & model-based (ours)\\
    
   \cline{3-10}
    \multicolumn{2}{|c||}{} & centralized & \multicolumn{2}{c||}{CTDE} & \multicolumn{4}{c||}{independent learners} & CTDE \\ 
    \cline{3-10}
    \multicolumn{2}{|c||}{} &  IQL  &  MAIQL & MATD3+BC &  IBC  &  ITD3+BC  &  ICQL  &  IOMAR  &  \textbf{MOMA-PPO} \\
    \hline
\multirow{1}{*}{FO} & all-observant &  \textbf{1.07 $\pm$ 0.01}  &  0.96 $\pm$ 0.05 & 1.04 $\pm$ 0.01 &  1.02 $\pm$ 0.01  &  0.78 $\pm$ 0.00  &  0.48 $\pm$ 0.06  &  0.73 $\pm$ 0.01  &  \textbf{1.07 $\pm$ 0.01}  \\ 
\hline
\multirow{2}{*}{PO} & independent & &  \textbf{0.92 $\pm$ 0.04}  &  0.59 $\pm$ 0.03 &  0.76 $\pm$ 0.04  &  0.30 $\pm$ 0.11  &  0.46 $\pm$ 0.04  &  0.45 $\pm$ 0.02  &  \textbf{0.95 $\pm$ 0.06}  \\ 
& leader-only & &  0.80 $\pm$ 0.05  &  0.73 $\pm$ 0.02 & 0.84 $\pm$ 0.02  &  0.48 $\pm$ 0.04  &  0.31 $\pm$ 0.05  &  0.39 $\pm$ 0.02  & \textbf{ 1.00 $\pm$ 0.01}  \\
\hline
    \end{tabular}}
     \vspace{6pt}\\
    
    (b) Four-agent Ant, various datasets and full/partial observability\\

    \resizebox{.98\textwidth}{!}{\begin{tabular}{|c|c||c||c|c||c|c|c|c||c|}
    \hline
    \multicolumn{2}{|c||}{\multirow{3}{*}{\diagbox{Tasks}{Algorithms}}} & \multicolumn{7}{c||}{model-free} & model-based (ours)\\
    \cline{3-10}
    \multicolumn{2}{|c||}{} & centralized & \multicolumn{2}{c||}{CTDE} & \multicolumn{4}{c||}{independent learners} & CTDE \\ 
    \cline{3-10}
    \multicolumn{2}{|c||}{} &  IQL  &  MAIQL &  MAIQL-ft  &  IBC  &  ITD3+BC  &  ICQL  &  IOMAR  &  \textbf{MOMA-PPO} \\
    \hline
\multirow{4}{*}{FO} & ant-random &  0.12 $\pm$ 0.00  &  0.28 $\pm$ 0.01  &  0.28 $\pm$ 0.03  &  0.31 $\pm$ 0.00  &  0.22 $\pm$ 0.02  &  0.08 $\pm$ 0.00  &  0.08 $\pm$ 0.00  &  \textbf{0.52 $\pm$ 0.07}  \\ 
& ant-medium &  0.97 $\pm$ 0.02  &  0.85 $\pm$ 0.02  &  0.81 $\pm$ 0.02  &  0.84 $\pm$ 0.01  &  1.04 $\pm$ 0.00  &  0.88 $\pm$ 0.12  &  1.10 $\pm$ 0.03  &  \textbf{1.29 $\pm$ 0.06}  \\ 
& ant-full-replay &  1.22 $\pm$ 0.02  &  0.77 $\pm$ 0.21  &  0.95 $\pm$ 0.13  &  1.20 $\pm$ 0.01  &  1.33 $\pm$ 0.01  &  1.21 $\pm$ 0.02  &  1.30 $\pm$ 0.00  &  \textbf{1.42 $\pm$ 0.07}  \\ 
& ant-expert  &  1.26 $\pm$ 0.01  &  1.24 $\pm$ 0.00  &  1.06 $\pm$ 0.07  &  1.24 $\pm$ 0.00  &  1.25 $\pm$ 0.02  &  0.73 $\pm$ 0.15  &  1.16 $\pm$ 0.01  &  \textbf{1.49 $\pm$ 0.01}  \\ 
\hline
\multirow{4}{*}{PO} & ant-random &                               &  0.31 $\pm$ 0.00  &  0.34 $\pm$ 0.04  &  0.31 $\pm$ 0.00  &  0.31 $\pm$ 0.00  &  0.17 $\pm$ 0.02  &  0.21 $\pm$ 0.02  &  \textbf{0.42 $\pm$ 0.05}  \\ 
& ant-medium  &                               &  0.14 $\pm$ 0.02  &  0.11 $\pm$ 0.01  &  0.17 $\pm$ 0.01  &  0.22 $\pm$ 0.05  &  0.09 $\pm$ 0.02  &  0.06 $\pm$ 0.01  &  \textbf{0.54 $\pm$ 0.19}  \\ 
& ant-full-replay &                               &  0.18 $\pm$ 0.02  &  -0.07 $\pm$ 0.10  &  0.21 $\pm$ 0.02  &  0.20 $\pm$ 0.01  &  0.09 $\pm$ 0.01  &  0.11 $\pm$ 0.02  &  \textbf{0.46 $\pm$ 0.10}  \\ 
& ant-expert &                               &  -0.16 $\pm$ 0.01  &  -0.23 $\pm$ 0.02  &  0.05 $\pm$ 0.04  &  0.16 $\pm$ 0.00  &  0.11 $\pm$ 0.03  &  0.10 $\pm$ 0.01  &  \textbf{0.18 $\pm$ 0.00} \\
\hline
    \end{tabular}}
\end{tabular}
\end{table*}
The experimental procedure such as hyperparameters, training routine, and raw learning curves are detailed in Appendix~\ref{app:results}. All algorithms are trained to convergence and we used 10 seeds for the Iterated Coordination Game and 3 seeds for MAMuJoCo tasks. Tables are normalized and report the mean evaluation performance and the standard error of the mean across seeds. Evaluation is done for 100 episodes using the greedy policies (no sampling). 
\subsection{Strategy Agreement}

Table~\ref{tab:toy-results} reports the results for the offline Iterated Coordination Game and validates most of our intuitions about strategy agreement: the centralized execution (IQL) and model-based (MOMA-PPO) approaches are able to coordinate agents regardless of the datasets. On the other hand, independent BC agents imitate the dataset behavior and therefore coordinate only if the dataset majorly demonstrates coordination. Surprisingly, the CTDE model-free approach MAIQL is able to break symmetry and coordinate agents in the neutral dataset. We hypothesize that small numerical errors in the centralized value approximation have the team favor one equivalent strategy over the other. Unfortunately, the conservatism of model-free methods forces agents to stay close to the demonstrated behaviors and prevails over this brittle symmetry-breaking mechanism in the unfavorable -- i.e., uncoordinated -- dataset.

\textbf{ }\newline

Table~\ref{tab:reacher} (a) confirms that these insights on strategy agreement hold in the more complex two-agent Reacher environment: model-free methods struggle with strategy agreement, especially under partial observability\footnote{\url{https://sites.google.com/view/moma-ppo} shows independent learners converge to incompatible conventions.}. Again, CTDE methods -- particularly MAIQL -- tend to fare better than independent learners. Interestingly, IBC fares best among independent learners. Finally, our model-based CTDE approach \textit{MOMA-PPO solves strategy agreement} and performs on par with centralized execution (IQL), a setting that sidesteps the strategy agreement issue altogether. MOMA-PPO matches or significantly outperforms all other baselines.

\subsection{Strategy Fine-Tuning}
From Table~\ref{tab:reacher} (b) one can investigate how the different offline methods cope with strategy finetuning. First, IQL performs on par with the other model-free methods which suggests that centralized execution (single-agent) vs. decentralized execution (multi-agent) is less a consideration for strategy fine-tuning than it is for strategy agreement. Yet, this also highlights that \textit{model-free methods (even when centralized) are unable to perform strategy fine-tuning.} Indeed, they are surpassed by our model-based method, MOMA-PPO. This latter generates additional synthetic experiences that \emph{allow for strategy fine-tuning in addition to strategy agreement}.
For model-free methods, independent learners tend to outperform CTDE ones (which echoes \cite{pan2022plan} observations). Additionally, comparing IQL with MAIQL performance does highlight that offline multi-agent coordination is more challenging in varied datasets (i.e. medium or full-replay that display both coordinated and uncoordinated behaviors) than in uniformly coordinated datasets (i.e. expert).

In partially observable (PO) tasks, model-free methods are unable to adapt the behaviors demonstrated in the datasets and they result in teams that run in circles because the yellow agent (the only one to observe the torso's headings and velocities) fails to correct the other limbs' motions. Conversely, with MOMA-PPO, the yellow agent steers the ant toward the correct direction, and the teams reach very satisfactory performances provided that the datasets have enough coverage to learn a world model that can simulate diverse and robust behaviors (cf. the lower performance for the expert dataset).\footnote{\url{https://sites.google.com/view/moma-ppo} shows rollouts with and without ``steering" behavior.} 

Finally, the poor performance of MAIQL-ft suggests that IQL's finetuning abilities might not carry over to the multi-agent setting (even though we used the ground-truth simulator to generate the rollouts). While there might be multiple causes, we hypothesize that it is mainly due to MAIQL's instability since it required intensive hyperparameters finetuning and filtering out collapsed runs for ant tasks. We believe that MAIQL's instability is exacerbated by the induced non-stationarity of the training data when augmenting it with online interactions. Unlike online methods, offline algorithms are designed to learn on fixed datasets and are thus ill-equipped to deal with data continually collected by changing policies. Similarly, experiments that used MAIQL instead of MAPPO for MOMA (i.e. MOMA-IQL) quickly led to unstable learning and exploding losses.

In conclusion, our results show that current model-free offline MARL methods fail at offline coordination. Crucially, this deficiency remains even in very simple domains such as the Iterated Coordination Game. Also, it sometimes leads to underperforming basic imitation learning (i.e., IBC) or the datasets' average score (see Table~\ref{tab:dataset_metrics}). Conversely, our model-based approach is able to coordinate agents even under severe partial observability and with learned world models by restoring inter-agent interactions throughout learning.

\subsection{Ablations} 
\label{sec:ablations}
\begin{figure}[h]
    \centering
    \vspace{-11pt}
    \begin{tabular}{c}
        \includegraphics[width=0.45\textwidth]{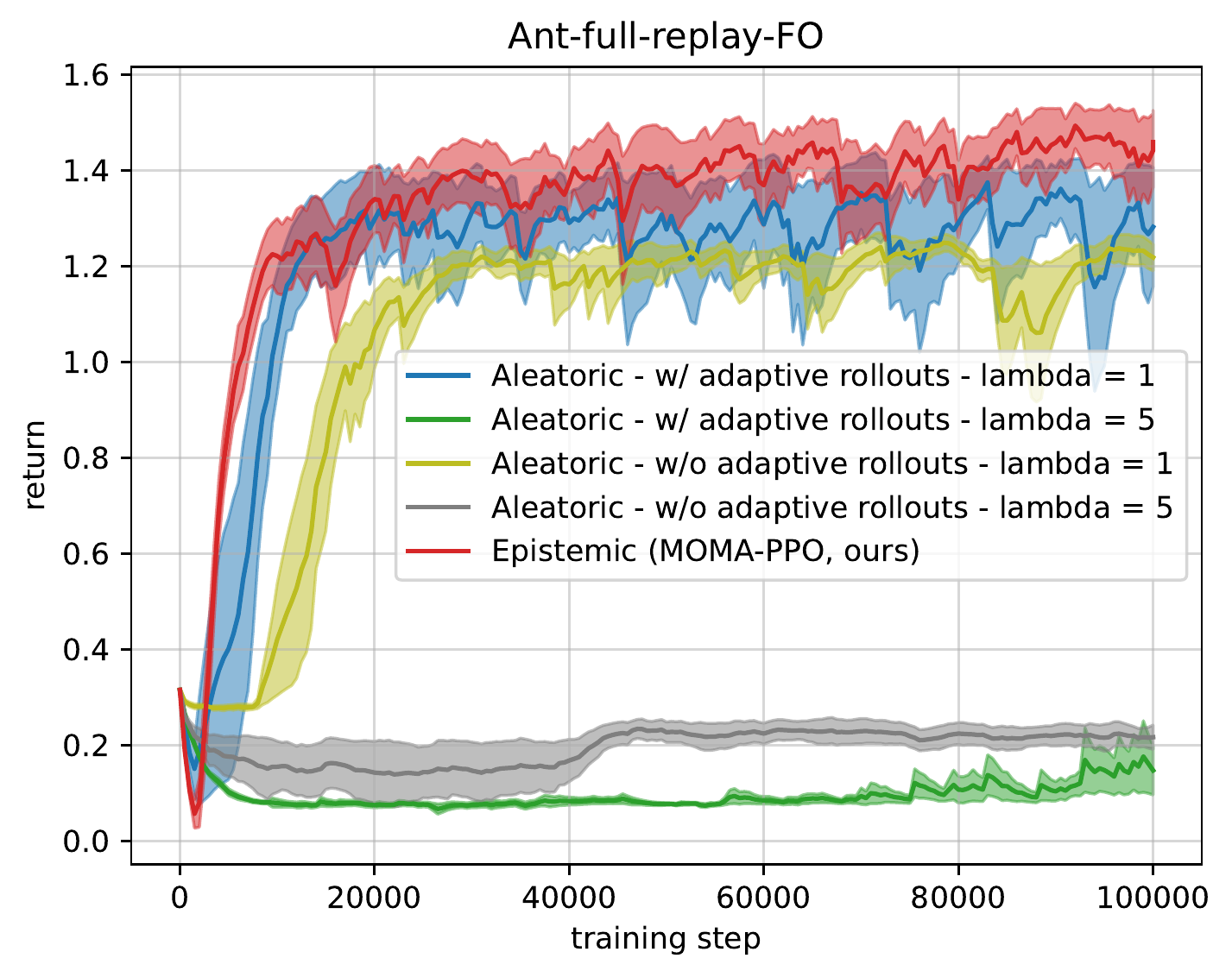}\\
        \includegraphics[width=0.45\textwidth]{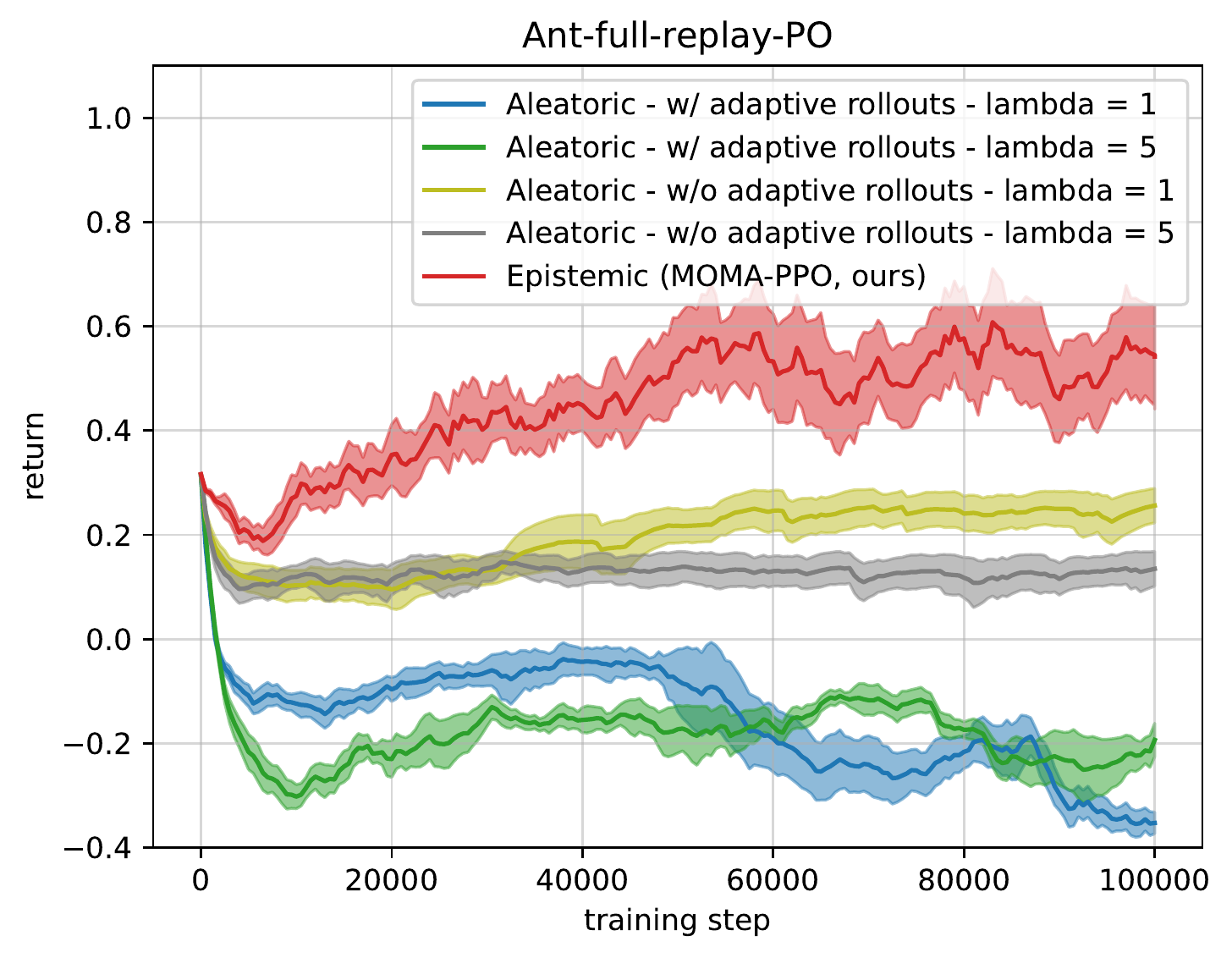}
    \end{tabular}
    \vspace{-12pt}
    \caption{Comparision between using epistemic uncertainty reward penalty (MOMA-PPO) vs. aleatoric uncertainty reward penalty (MOPO-like). Mean and standard error of the mean on three seeds.}
    \label{fig:ablations-epistemic}
    \vspace{-6pt}
\end{figure}

We validate our design choice of using epistemic uncertainty over aleatoric. To do so we replace MOMA-PPO's penalty with the aleatoric uncertainty penalty of MOPO \cite{yu2020mopo}. For a fair comparison, we followed \cite{yu2020mopo}'s parameter search procedure (i.e., coefficient values of 1 and 5). Additionally, we investigated aleatoric penalty both with and without the use of adaptive rollout length (which is based on epistemic uncertainty). The resulting learning curves on Ant-full-replay with full and partial observability are displayed in Figure~\ref{fig:ablations-epistemic}. It appears that using epistemic uncertainty always significantly outperforms using aleatoric uncertainty except in the case of full observability and adaptive rollouts where the comparison is not statistically significant. This validates the soundness of our choice regarding the use of epistemic uncertainty over aleatoric uncertainty in deterministic environments.

Appendix~\ref{app:ablations} reports additional ablations, which main insights we discuss here. First, using the ground-truth simulator yielded higher scores suggesting that the performance of MOMA-PPO can be further improved by learning a more accurate world model. Then, not clipping the generated next-states to the dataset's bounding box prevented generating length 10 rollouts as the states' magnitude exploded after a few world-model steps. Also, the use of uncertainty penalty ($\lambda_g$) and adaptive rollouts' length ($l_\epsilon$) are necessary to achieve satisfactory results. Finally, varying the rollout's maximum length from 5 to 50 did not significantly impact performance.

\section{Discussion and Conclusion}
This work explores coordination in offline MARL and highlights the failures of current model-free methods. For instance, they struggle in the presence of multiple equivalent but incompatible optimal team strategies (\textit{strategy agreement}), or when partial observability requires the team to adapt the behaviors demonstrated in the dataset (\textit{strategy fine-tuning}). In such scenarios, prevalent model-free methods might even fail to match the dataset's performance and fall short compared to behavioral cloning. To address these problems, we propose MOMA-PPO which is, to our knowledge, the first model-based offline MARL approach. Our method is able to coordinate teams of offline learners and significantly outperforms model-free alternatives. Interestingly, it also outperforms the fully centralized model-free method IQL \cite{kostrikov2021offline} even though this latter completely bypasses the strategy agreement problem. This suggests that model-free methods, even when fully centralized, are unable to deal with strategy fine-tuning. We also observe in our experiments that the single-agent approach of fine-tuning offline methods (i.e. IQL) with online interactions \cite{kostrikov2021offline} fails in the multi-agent setting (i.e. MAIQL). Our model-based approach succeeds at both multi-agent strategy agreement and strategy fine-tuning problems which goes to show that the benefits of offline model-based approaches over offline model-free ones \cite{yu2020mopo} hold in the multi-agent setting. This might indicate that world models may generalize better than values networks in offline learning's limited data regime. Finally, for tasks that require adapting the team's behavior, dataset coverage might be more desirable than demonstrated performance. Indeed, methods fare better on random datasets than on expert ones. 

MOMA-PPO's successes put forward the benefits of model-based methods that leverage online policy optimization. Nevertheless, our approach of coordinating agents through a world model is general and some might be interested in extending it to more sample-efficient policy learning algorithms (i.e., MOMA-SAC). Also, the dataset bounding box clipping and adaptive rollouts developed for MOMA-PPO might benefit the single-agent setting. 

Finally, this work aims to pinpoint an overlooked issue in the offline MARL community and proposes a new avenue of model-based solutions that shifts away from conventional model-free approaches. Therefore, we look forward to future work that will further analyze model-based offline MARL approaches and scale them to more domains and datasets. 


\begin{acks}
The authors would like to thank Luis Pineda, Hengyuan Hu, and Eugene Vinitsky for their help, insightful discussions, and advice. This research was enabled in part by support provided by Calcul Québec and the Digital Research Alliance of Canada. This work was partially conducted while Paul was interning at Meta AI - FAIR.
\end{acks}



\bibliographystyle{ACM-Reference-Format} 
\bibliography{AAMAS_2024_paper}


\newpage
\onecolumn
\appendix
\newcommand{\acrshort}[1]{\MakeUppercase{#1}}
\renewcommand{\citet}{\cite}

\section{Reproducibility details}
\label{app:method}

The following section focuses on reproducibility and goes into detail about the implementations and experimental procedures.
\subsection{Methods implementation}
\label{sec:ocp-method-details}
\paragraph{\textbf{\acrshort{moma-ppo} entropy bonus and action penalty.}}
 For offline methods based on online \acrshort{rl}  algorithm, exploration is an important component (cf. \acrshort{td3}'s exploration strategy) so we used an entropy bonus for \acrshort{ppo} defined as:
\begin{python}
# entropy bonus
# dimensions are batch, act_dim, n_agents. 
# Instead of using closed form entropy, 
# we estimate it with E(-pi log pi) were
# the expectation is sampled over pi_old 
# so we have to correct with pi_new/pi_old (which is ratio!)

# we do two losses 
# (clipped / not clipped like with the actor loss)
surrogate_entropy = - (ratio * new_policy).mean(0)
clipped_entropy = - (clipped_ratio * new_policy).mean(0)

entropy = torch.min(surrogate_entropy, clipped_entropy)

self.entropy_alpha += self.ppo_entropy_bonus_coeff*(self.ppo_entropy_target - entropy).detach()
self.entropy_alpha.data.clamp_min_(0.)

# minus sign because we minimize the expressions 
# i.e. max(ent) = min(-ent)
entropy_bonus = - entropy * self.entropy_alpha
\end{python}
with an entropy bonus coefficient of 0.001 and an entropy target of -6 and -4 for respectively Ant and Reacher tasks. Since entropy computation can become numerically unstable for squashed actions close to the Tanh bounds, \acrshort{ppo} uses action penalty instead of Tanh squashing to keep actions close to the -1, 1 range: 
\begin{python} 
delta = (1. - actions.abs())
action_bound_error = ((delta < 0.).to(torch.float32) * delta**2).sum(1, keepdim=True)

surrogate_action_bound_error = (ratio * action_bound_error).mean(0)
clipped_action_bound_error = (clipped_ratio * action_bound_error).mean(0)

action_bound_error = torch.max(surrogate_action_bound_error, clipped_action_bound_error)
action_penalty = self.ppo_action_penalty_coeff * action_bound_error
\end{python}
with an action penalty coefficient of 1.

\paragraph{\textbf{General Advantage Estimation.}}

We show below how we modified the general advantage estimation to account for rollout termination (indicated by \verb!time_out_masks!).
\begin{python}
# initial (end) running returns is the next state value
running_returns = next_values[-1] * masks[-1]

# initial (end) advantage is 0 because no difference in value and return
running_advants = 0

for t in reversed(range(0, len(rewards))):

    # We are going reverse so if done, only reward because end of 
    # episode if timeout, we stop accumulation and use value as 
    # bootstrap like in initialization 
    running_returns = rewards[t] + self.discount * masks[t] * (running_returns * time_out_masks[t] + (1. - time_out_masks[t]) * next_values[t] * masks[t])
    
    returns[t] = running_returns

    ## No accumulation here and timeout doesn't influence next_state value
    running_delta = rewards[t] + (self.discount * next_values[t] * masks[t]) - values[t]

    ## if timeout running_advants goes back to running_delta because
    # we do not have extra rewards to estimate it 
    # (cf initialization above)
    running_advants = running_delta + (self.discount * self.lamda * running_advants * masks[t]) * time_out_masks[t]

    advants[t] = running_advants
\end{python}

\paragraph{\textbf{Memory module for partial observability.}}
To handle partial observability we use observation-action histories $h^i_t$ of size ten (i.e. the ten past observation-action pairs). These histories are processed with self-attention followed by soft-attention to yield embeddings of size $e_h=128$ that are concatenated to the current state before being fed to the policy and value networks. We use a first linear layer to encode $h^i_t$ to $\R^{e_h}$ and add positional encodings \cite{vaswani2017attention}. Query, Key, and Value networks are linear layers and we follow \citet{vaswani2017attention}'s Scaled Dot-Product Attention with skip connection and layer-norm. Finally, the resulting self-attentions are aggregated using soft-attention with the soft-key network being a bias-less linear layer and the soft-queries are $e_h$ normally initialized trainable parameters.

Note that policy and value networks use (and backprop through) the same memory module but target networks have their own target memory module (that tracks the memory module with Polyak updates just like regular target networks). The memory learning rate is $1\e{-4}$ for all the algorithms.

\paragraph{\textbf{MAIQL.}}
We provide IQL \cite{kostrikov2021offline} learning rules here for completeness.
Value learning is done with SARSA Bellman on $e$ expectile of $Q$ instead of mean $Q$ (this latter is a special case where $e=0.5$). To ensure that the expectile is computed only from the action selection distribution and is not influenced by the randomness of the environment's transitions, IQL uses a state-only value function $V$ that marginalizes over future transitions:
\begin{equation}
\label{eq:LV}
\begin{split}
L_V(\psi)&= \E_{s,a\sim \gD}\left[\Lt(\Qtarget(s,a)-\V(s))\right]\\
&=\E_{s,a\sim \gD}\left[|e - \I\left(\Qtarget(s,a)-\V(s)<0\right)|\left(\Qtarget(s,a)-\V(s)\right)^2\right],\\
\end{split}
\end{equation}
\begin{equation}
\label{eq:LQ}
L_Q(\theta)=\E_{s,a,r,s'\sim\gD}\left[\left(r(s,a)+\gamma\V(s')-\Q(s,a) \right)^2\right].
\end{equation}

Policy extraction is done with
Advantage Weighted Regression (AWR):
\begin{equation}
\label{eq:Lpi}
L_\pi(\phi)=\E_{(s,a)\sim\gD}\left[-\exp\left(\beta\left(\Qtarget(s,a)-\V(s)\right)\right)\log \policy(a|s)\right].
\end{equation}

We make IQL multi-agent (i.e. MAIQL) by leveraging the CTDE formalism and  using QMIX value-decomposition \citep{rashid2018qmix} for both $Q$ and $V$: 
\begin{equation}
\begin{split}
    \V(s)=\sum_i\wiv(s)\Vi(h^i) + \bv(s),\\
\Q(s)=\sum_i\wiq(s)\Qi(h^i) + \bq(s).
\end{split}
\end{equation}
And the target network:
\begin{equation}
  \Qtarget(s)=\sum_i\wiqtarget(s)\Qitarget(s^i) + \bqtarget(s).
\end{equation}

Similarly, the joint policy is assumed to decompose as: 
\begin{equation}
\label{eq:pimix}
    \pi(a|s) \triangleq \prod_i\pi_i(a^i|h^i).
\end{equation}

Injecting this into \eq\ref{eq:Lpi} one gets:
\begin{equation}
\begin{split}
L_{\pi}(\phi)&=
\E_{(s,a)\sim\gD}\left[-B(s)
\prod_i\exp\left(\beta(\wiqtarget(s)\Qitarget(h^i,a^i)-\wiv(s)\Vi(h^i))\right)
\sum_j\log\pi_{\phi^j}(a^j|h^j)\right],\\
B(s)&=\exp(\bqtarget(s)-\bv(s)).
\end{split}
\end{equation}

\paragraph{\textbf{Finetuning \acrshort{maiql}: \acrshort{maiql}-ft.}}
We follow \citet{kostrikov2021offline}'s finetuning procedure and use the ground truth simulator to generate the rollout. However, the rollouts are still generated using MOMA's Dyna-like approach described in Section~\ref{sec:method} rather than generating length 1000 rollouts from the initial state distribution. Indeed, we consider the model-based offline setting and not the offline-to-online finetuning setting. Therefore it is unfeasible to assume that the task's ground-truth initial state distribution is known or that it is possible to learn a world model that remains accurate over 1000 simulation steps.

\paragraph{\textbf{Opensource baselines.}}
For the baseline implementations, we followed the official repositories at:
\begin{itemize}[noitemsep]
    \item \url{https://github.com/sfujim/TD3_BC},
    \item \url{https://github.com/ling-pan/OMAR/},
    \item and \url{https://github.com/ikostrikov/implicit_q_learning}.
\end{itemize}

\clearpage
\subsection{Hyperparameters, Tuning, and Training}

Unless specified otherwise, networks are two-layers 256 units ReLu MLPs. We use Adam optimizer \cite{kingma2014adam} with default hyperparameters (except learning rates) and a batch size of 256. We clip gradient norms to 1.

\paragraph{\textbf{World model learning.}}
World models use four layers and 1024 units.  World models use a learning rate of $3\e{-5}$ and are trained for $3\e{6}$ steps. 

\paragraph{\textbf{Policy learning.}}
\acrshort{mappo} learning rate was finetuned on \acrshort{mamujoco} online task \\ \verb!halfcheetah-v2_2x3_full! with a grid search across [$1\e{-6}, 5\e{-6}, 1\e{-5}, 5\e{-5}, 1\e{-4}, 5\e{-4}, 1\e{-3}$]. We kept the value of $5\e{-5}$ for \acrshort{moma-ppo} for all experiments. \acrshort{ppo} uses rollouts length of 1000, 5 epochs per update, 2000 transitions between updates, a clip value of 0.2, a $\lambda$ value of 0.98, and a critic loss coefficient of 0.5.

\acrshort{maiql} showed quite unstable and we had to finetune its learning rate extensively depending on the tasks. We tried the range [$1\e{-4}, 3\e{-4}, 1\e{-5}, 5\e{-5}, 1\e{-6}$] on ant-expert for \acrshort{maiql} and \acrshort{maiql}-ft and kept $3\e{-4}$. For the results on ant-expert with full observability, we had to discard one collapsed unstable seed for \acrshort{maiql} and retrain another seed. For reacher tasks we tried [$5\e{-5}, 1\e{-4}, 3\e{-4}, 5\e{-4}, 1\e{-3}$] and kept $3\e{-4}$. \acrshort{iql} and \acrshort{maiql} use an expectile value of 0.7 and an \acrshort{awr} temperature of 3.

All other methods use their default learning rate of $3\e{-4}$. \acrshort{itd3}+\acrshort{bc} uses a \acrshort{bc} regularization parameter of 2.5, a policy frequency update of 2, a policy noise of 0.2, and a noise clip value of 0.5. \acrshort{icql} uses a coefficient of 1 (and so does \acrshort{iomar}'s \acrshort{cql} component) for all tasks except for ant-expert tasks where we had to tune it between [$0.1, 0.5, 1, 5$] and kept a value of 5 (same for \acrshort{iomar}). Additionally, \acrshort{cql} uses an LSE temperature of 1, 10 sampled actions, and a sample noise level of 0.2. Finally, \acrshort{iomar} uses a coefficient of 1, 2 iterations, a $\mu$ value of 0, a $\sigma$ value of 2, 20 samples, and 5 elites. 

We use default values for other hyperparameters and we report them here for consistency. We use a discount factor of 0.99 and a target update coefficient of 0.005 for the Polyak averaging. \acrshort{td3} (and thus \acrshort{cql} and \acrshort{omar}) policies are deterministic with Tanh squashing while \acrshort{iql} and \acrshort{ppo} use Gaussians with state-dependent variance. \acrshort{iql} uses Tanh squashing while \acrshort{ppo} does not (see below).

Model-free methods are trained for $1\e{6}$ learning steps while \acrshort{moma-ppo} is trained for $1\e{5}$ learning steps which correspond to roughly $2\e{8}$ collected interactions from the world model. 

\subsection{Compute}
Our longest \acrshort{moma-ppo} training took 6 days on a Tesla P100-PCIE-12GB GPU. For comparison our longest \acrshort{iomar} run took 38 hours on a Tesla V100-SXM2-32GB GPU and our longest \acrshort{maiql} run took 40 hours on a Tesla V100-SXM2-32GB GPU and five days on a Tesla P100-PCIE-12GB GPU for \acrshort{maiql}-ft. Training a world model took at most three days on a Tesla P100-PCIE-12GB GPU. Note that training times are also impacted by how often we estimate performance for the learning curves and we do it much more often for \acrshort{moma-ppo} (5 episodes every 50 learning steps) than for other methods (10 episodes every 5000 learning steps). 

\clearpage
\section{Raw results}
\label{app:results}
This section focuses on transparency and provides the raw results of the experiments.
\subsection{Learning Curves}
\begin{figure}[H]
    \centering
    \begin{tabular}{c}
        \includegraphics[width=0.4\textwidth]{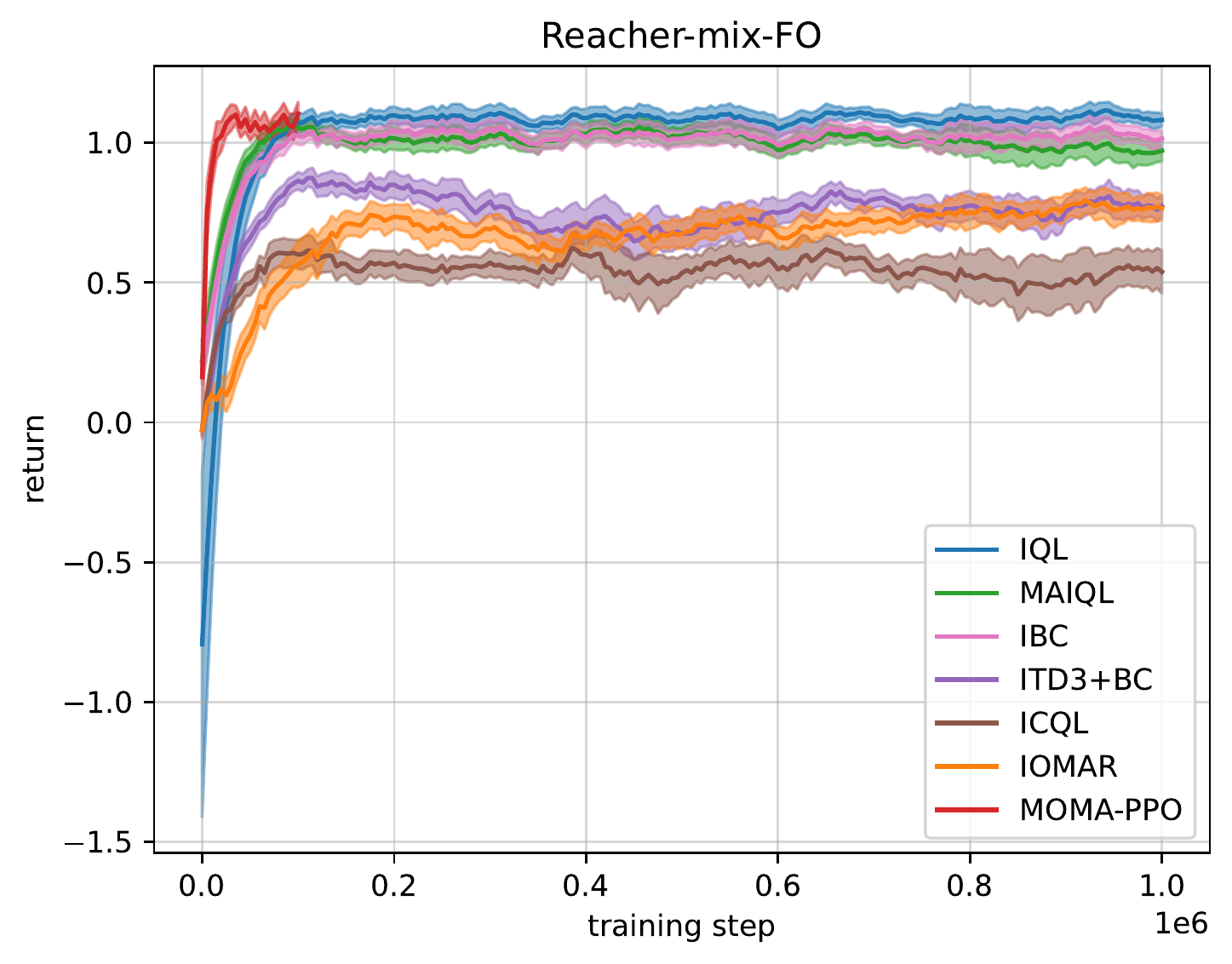}\\
        \includegraphics[width=0.4\textwidth]{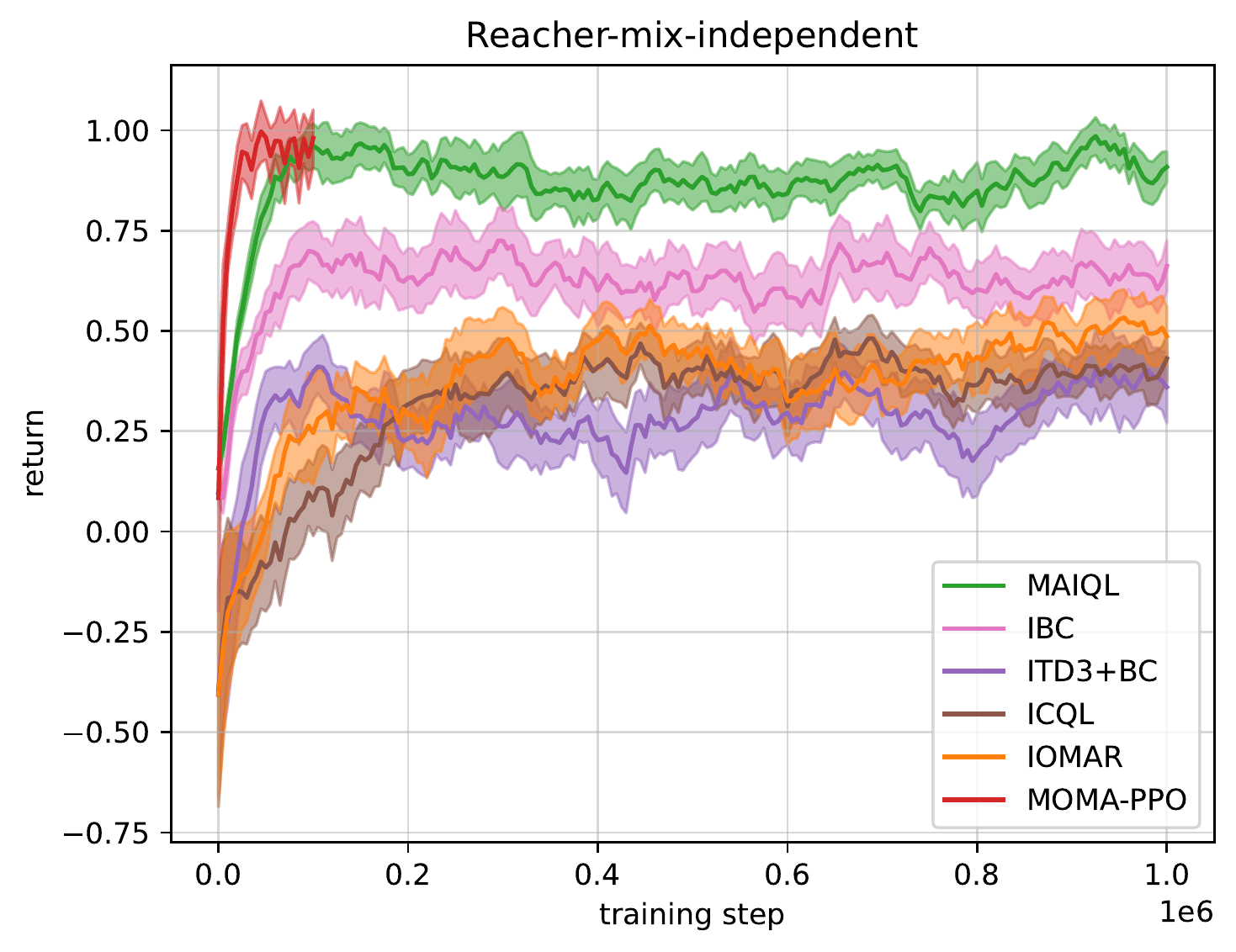} \\
        \includegraphics[width=0.4\textwidth]{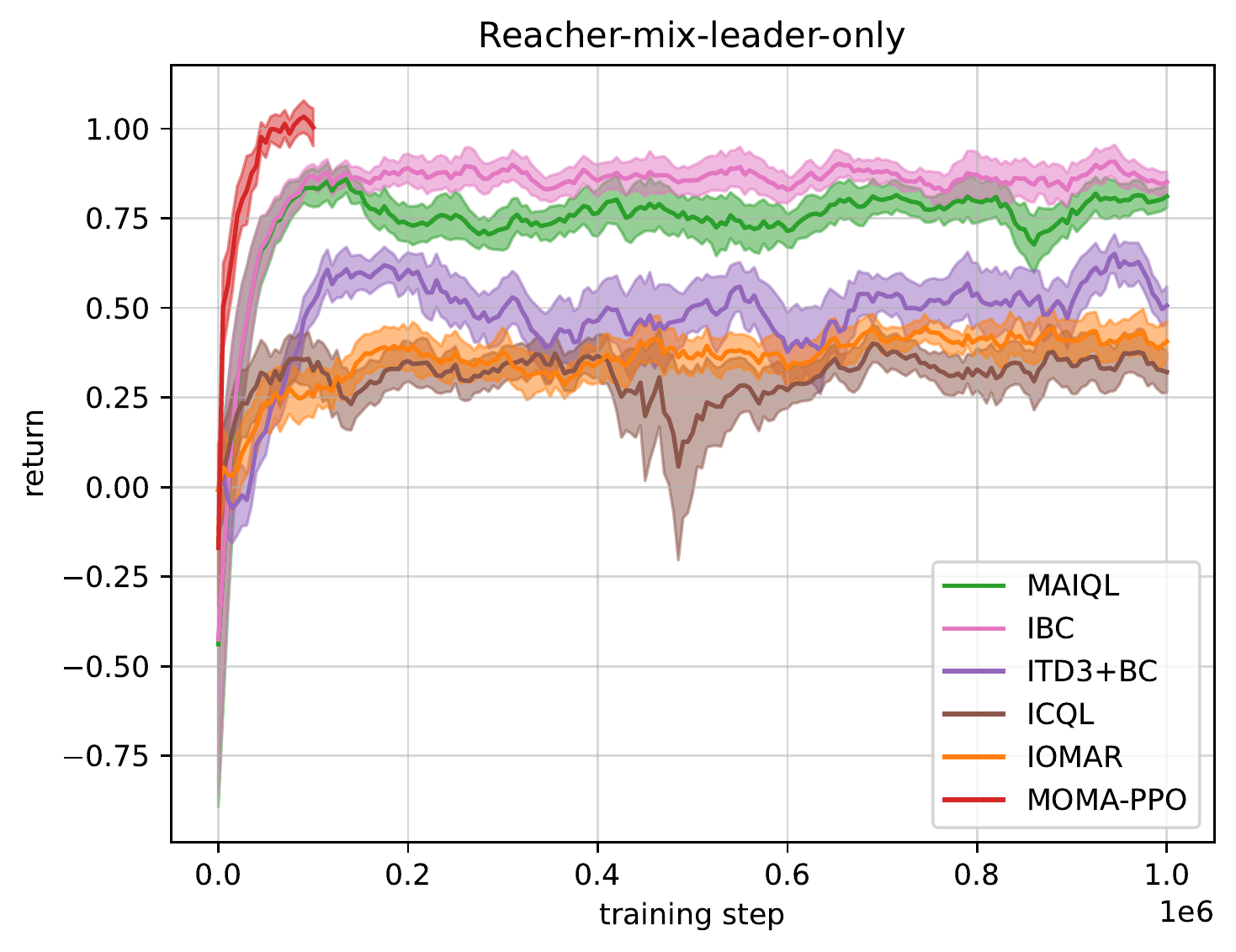}
    \end{tabular}
    \caption{Learning Curves for two-agent Reacher. Mean and standard error of the mean on three seeds.}
    \label{fig:learning-curves-reacher}
\end{figure}
\begin{figure}[H]
    \centering
    \begin{tabular}{cc}
         \includegraphics[width=0.35\textwidth]{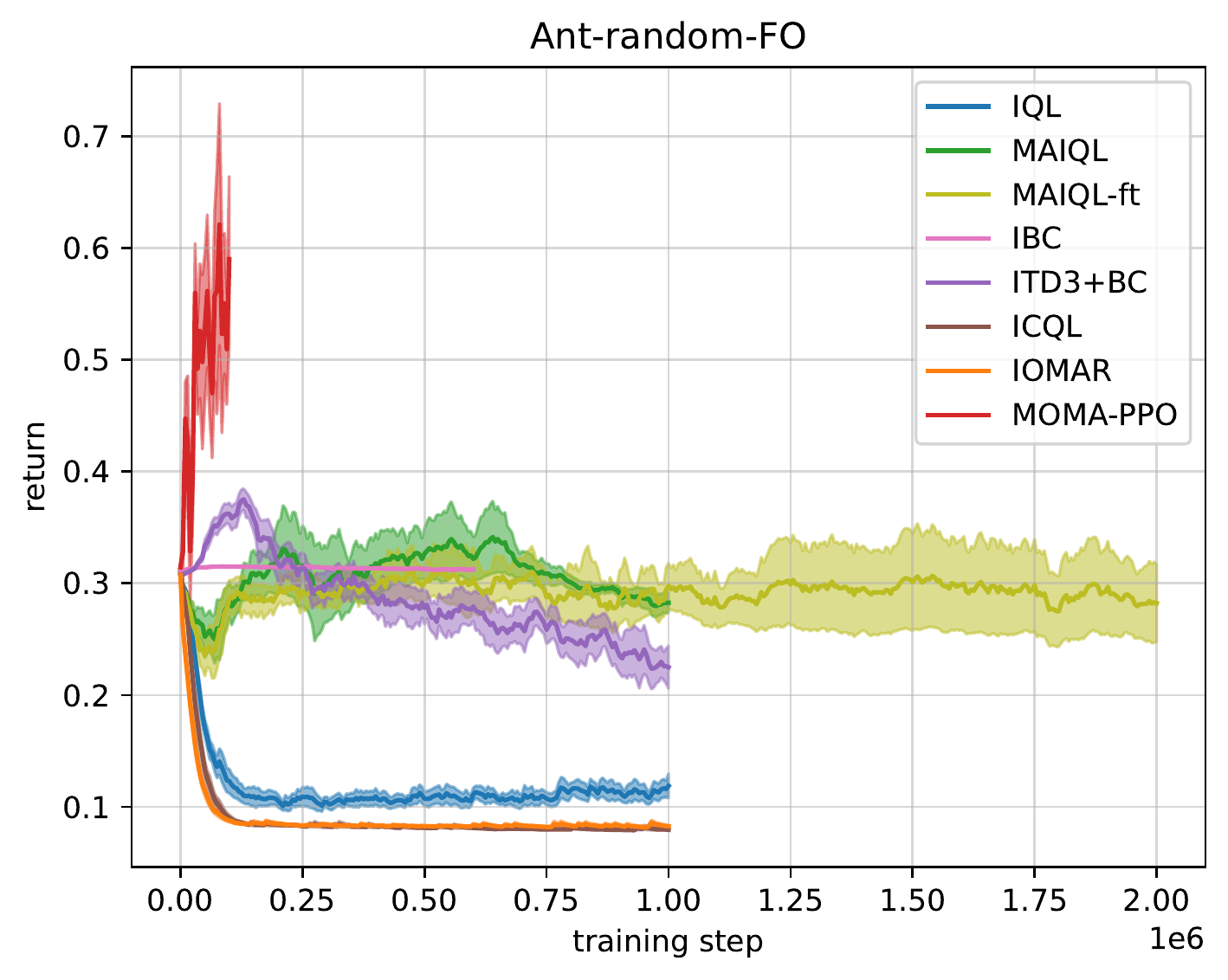} &
         \includegraphics[width=0.35\textwidth]{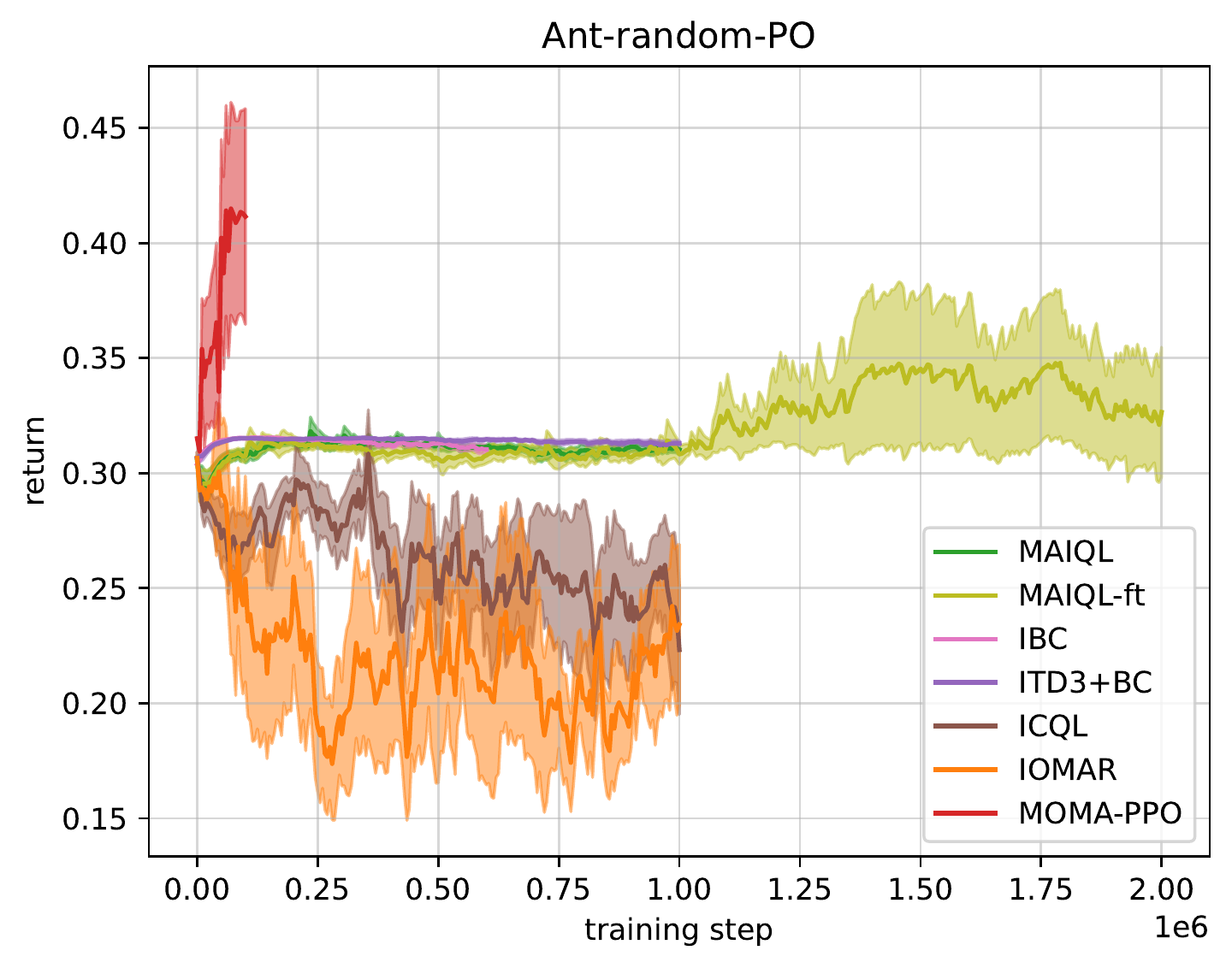}\\
         \includegraphics[width=0.35\textwidth]{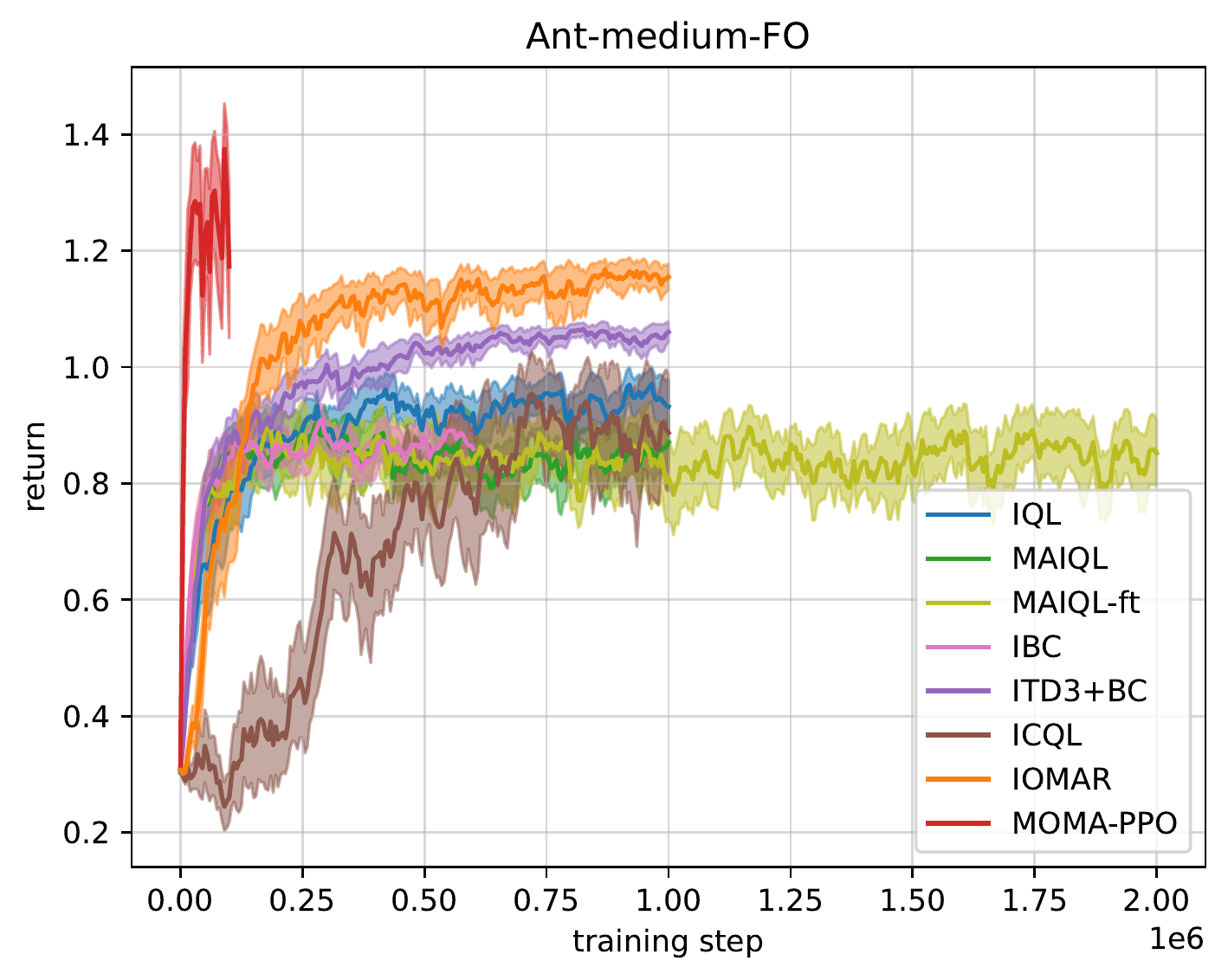} & 
         \includegraphics[width=0.35\textwidth]{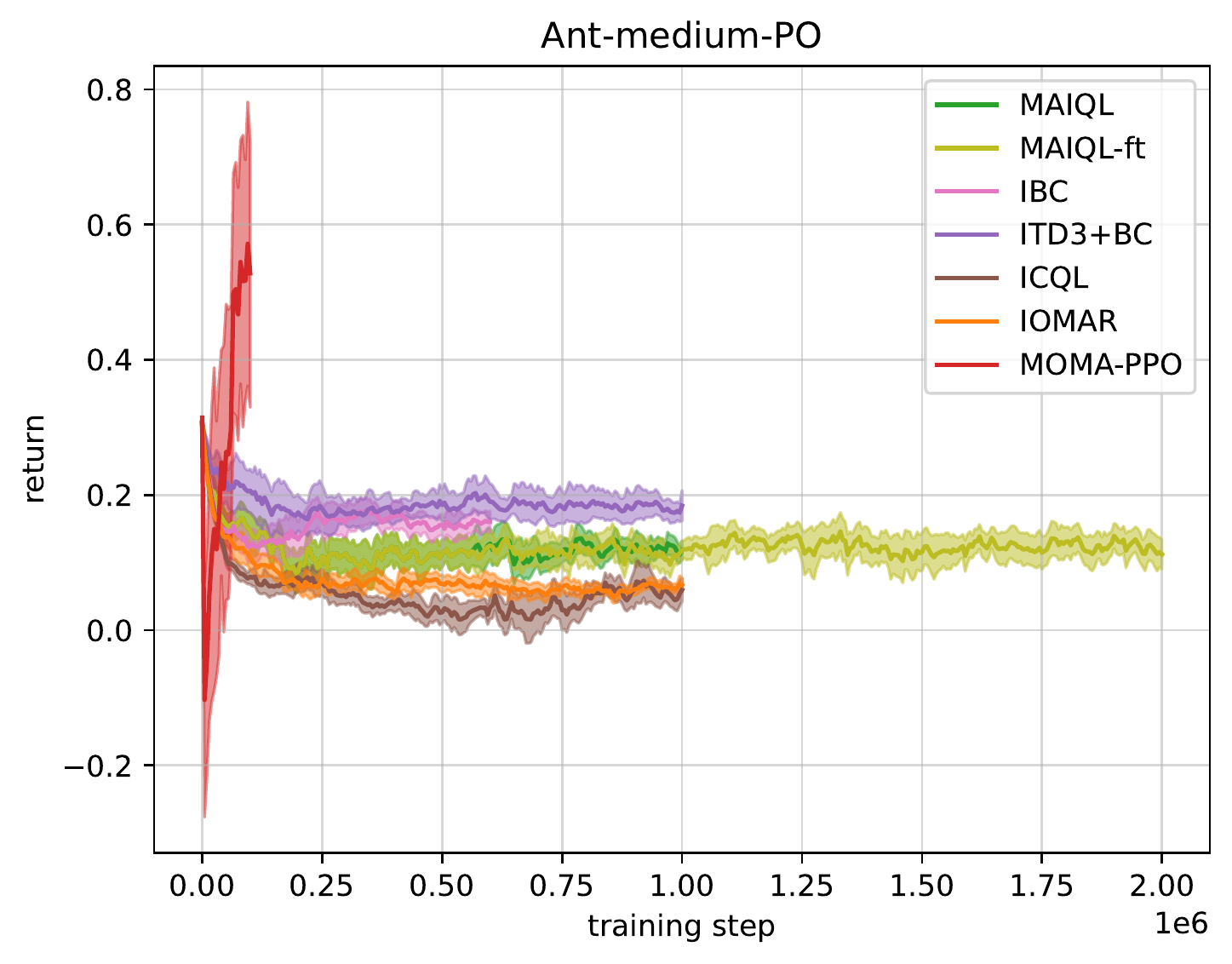}\\
         \includegraphics[width=0.35\textwidth]{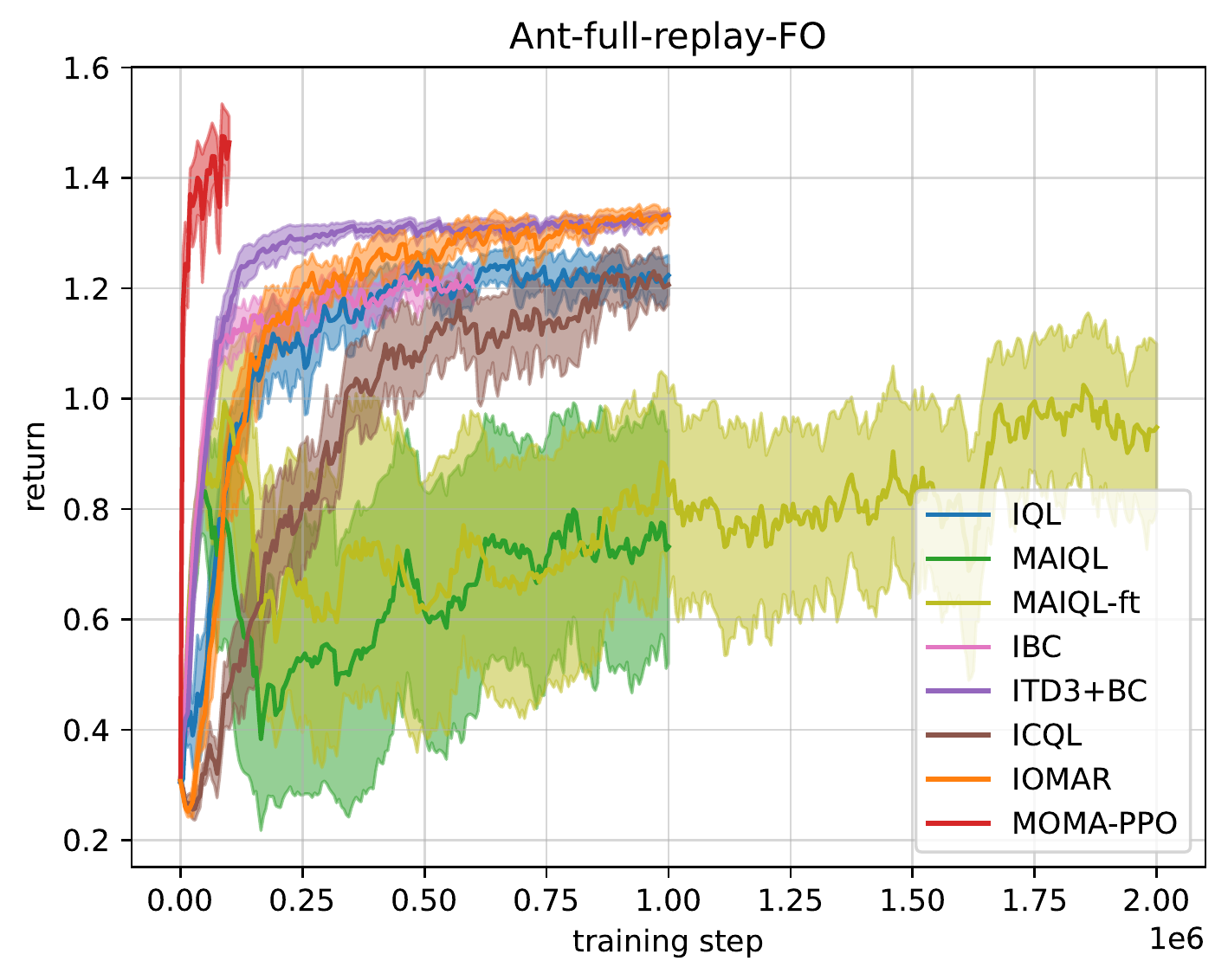} &
         \includegraphics[width=0.35\textwidth]{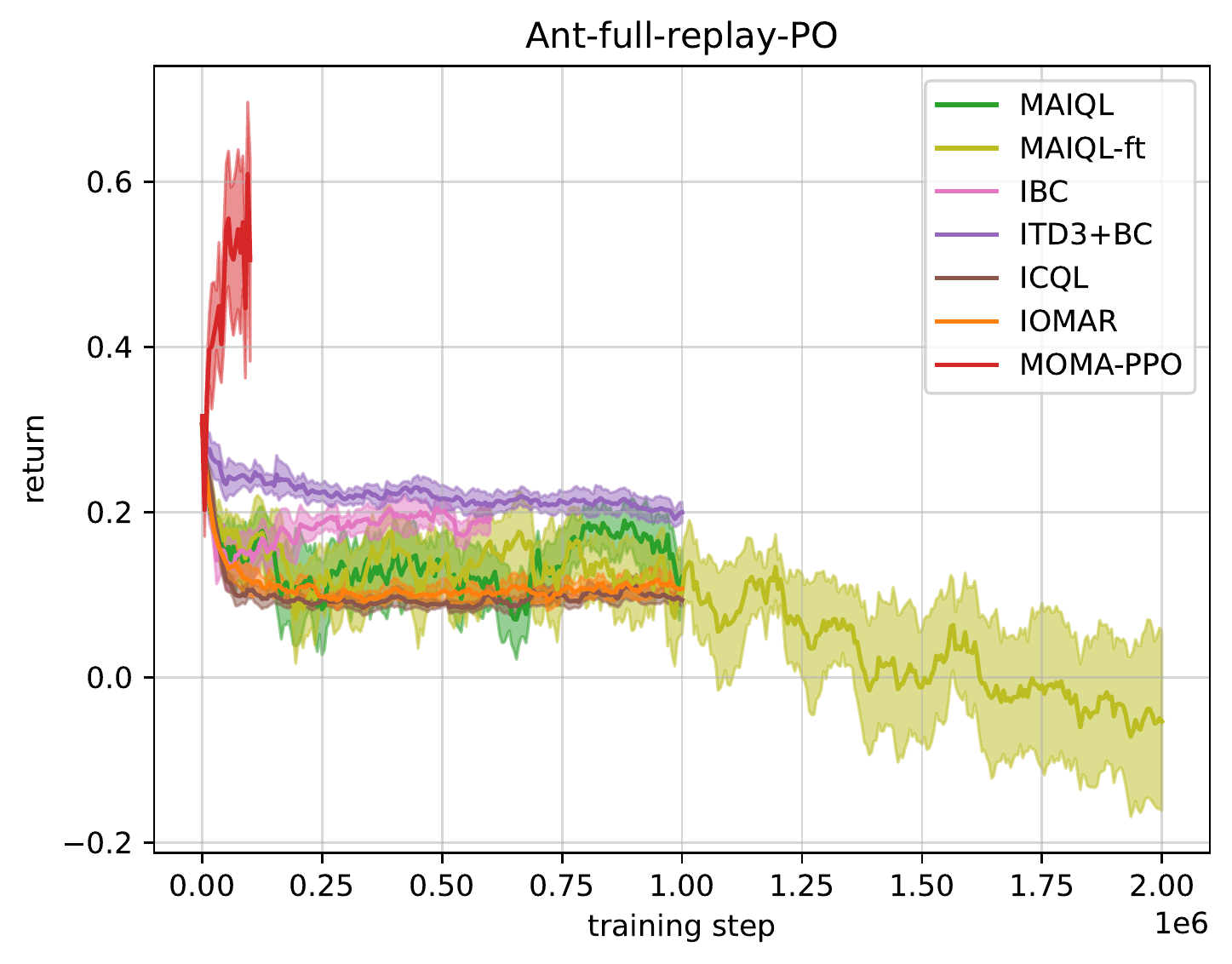}\\
         \includegraphics[width=0.35\textwidth]{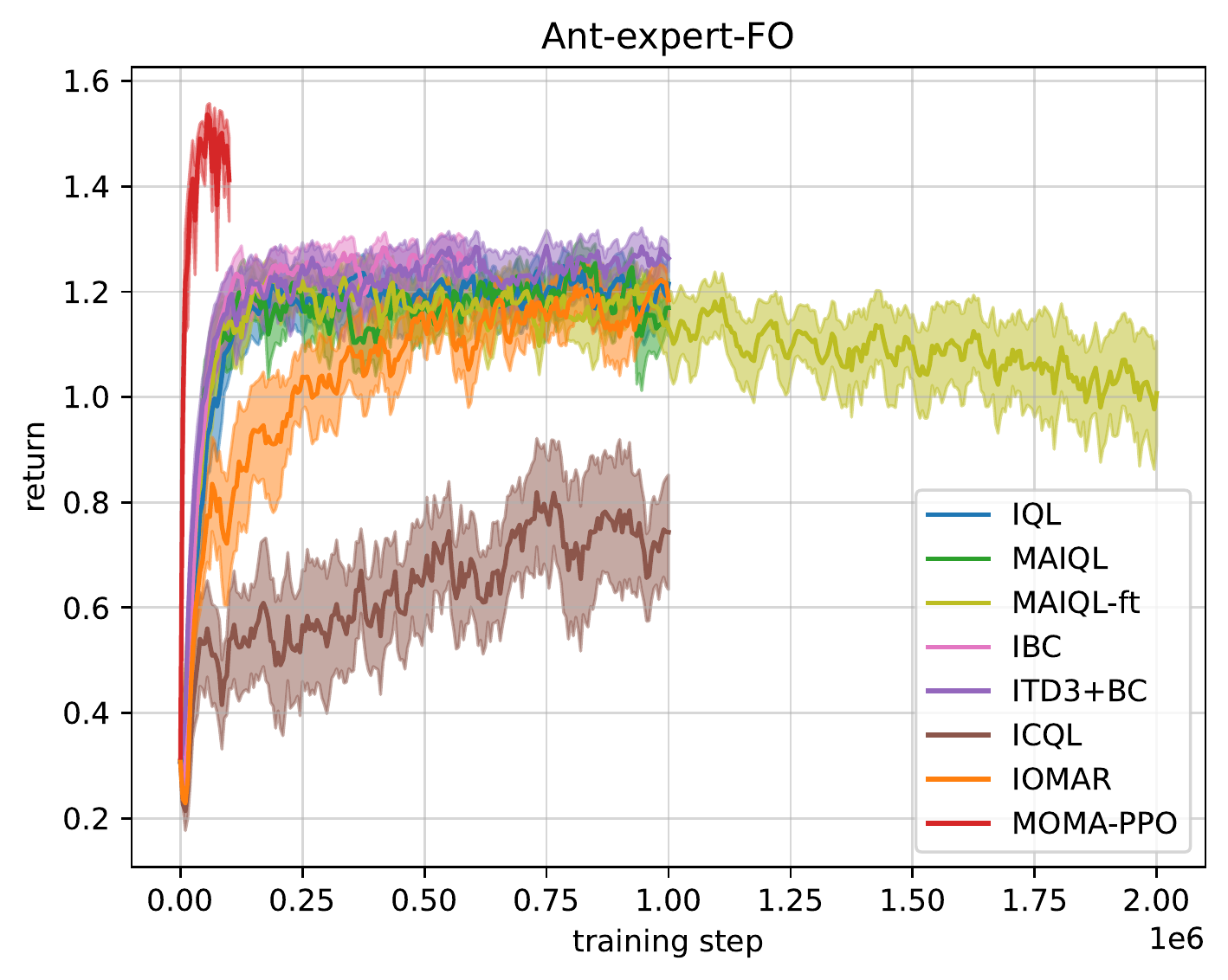} & 
         \includegraphics[width=0.35\textwidth]{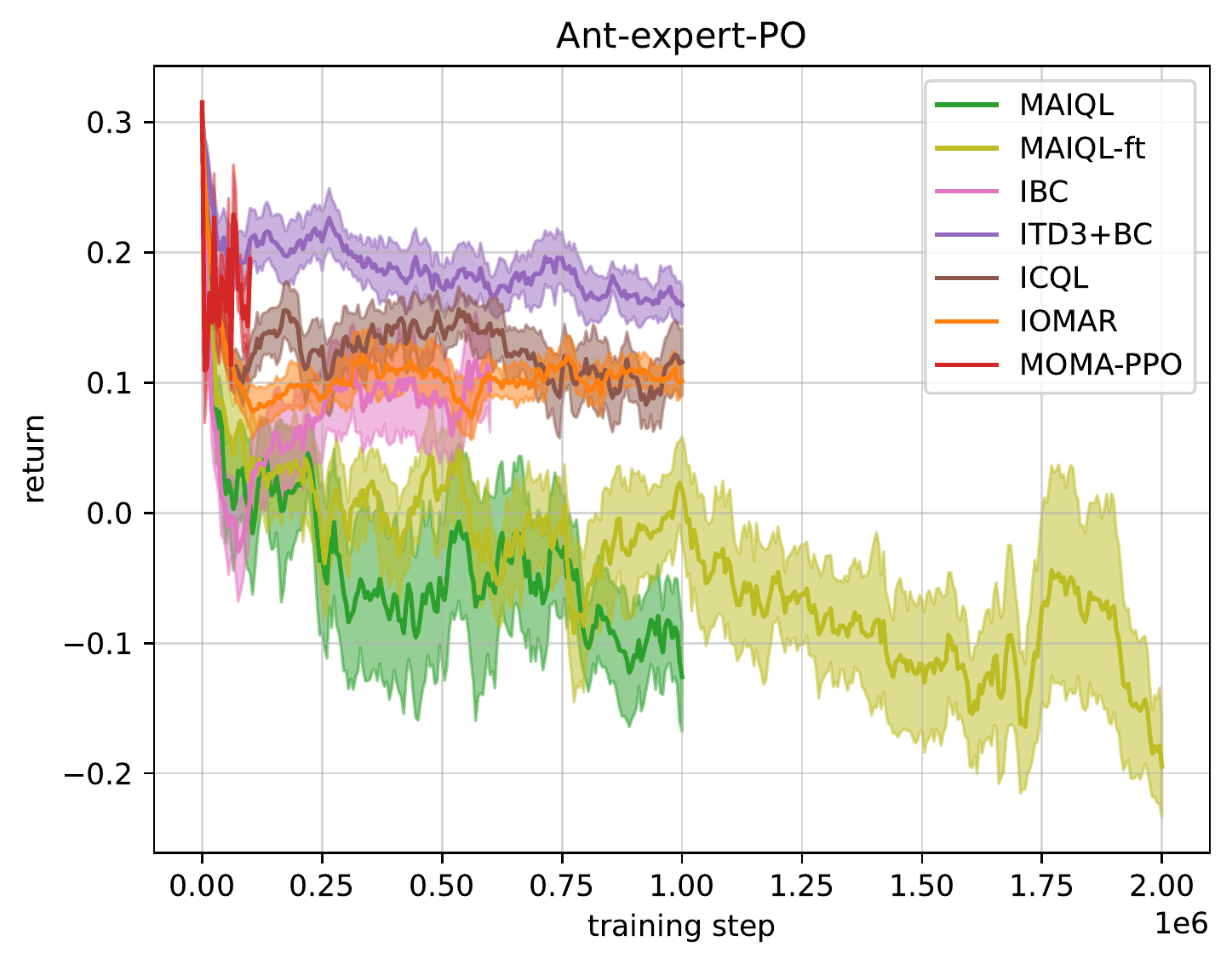}
    \end{tabular}
    \caption{Learning curves for four-agent Ant. Mean and standard error of the mean on three seeds.}
    \label{fig:learning-curves-ant}
\end{figure}

Figues~\ref{fig:learning-curves-reacher} and \ref{fig:learning-curves-ant} show the learning curves for the Reacher and Ant environments respectively. We use 5 episodes to evaluate MOMA-PPO every 50 learning steps and 10 episodes every 5000 learning steps to evaluate the other methods (this is why MOMA-PPO curves might look noisier). We used a smoothing factor of .8 for all curves. We report the mean over three seeds and the shaded area represents $\pm$ the standard error of the mean. We train MOMA-PPO for $1\e{5}$ training steps because it is on-policy while the off-policy methods are trained for $1\e{6}$ training steps. MAIQL-ft training is $2\e{6}$ steps (half offline and half online fine-tuning).

\subsection{Ablations}
\label{app:ablations}
\begin{figure}[H]
    \centering
    \resizebox{1\textwidth}{!}{
    \begin{tabular}{cc}
        \includegraphics[width=0.5\textwidth]{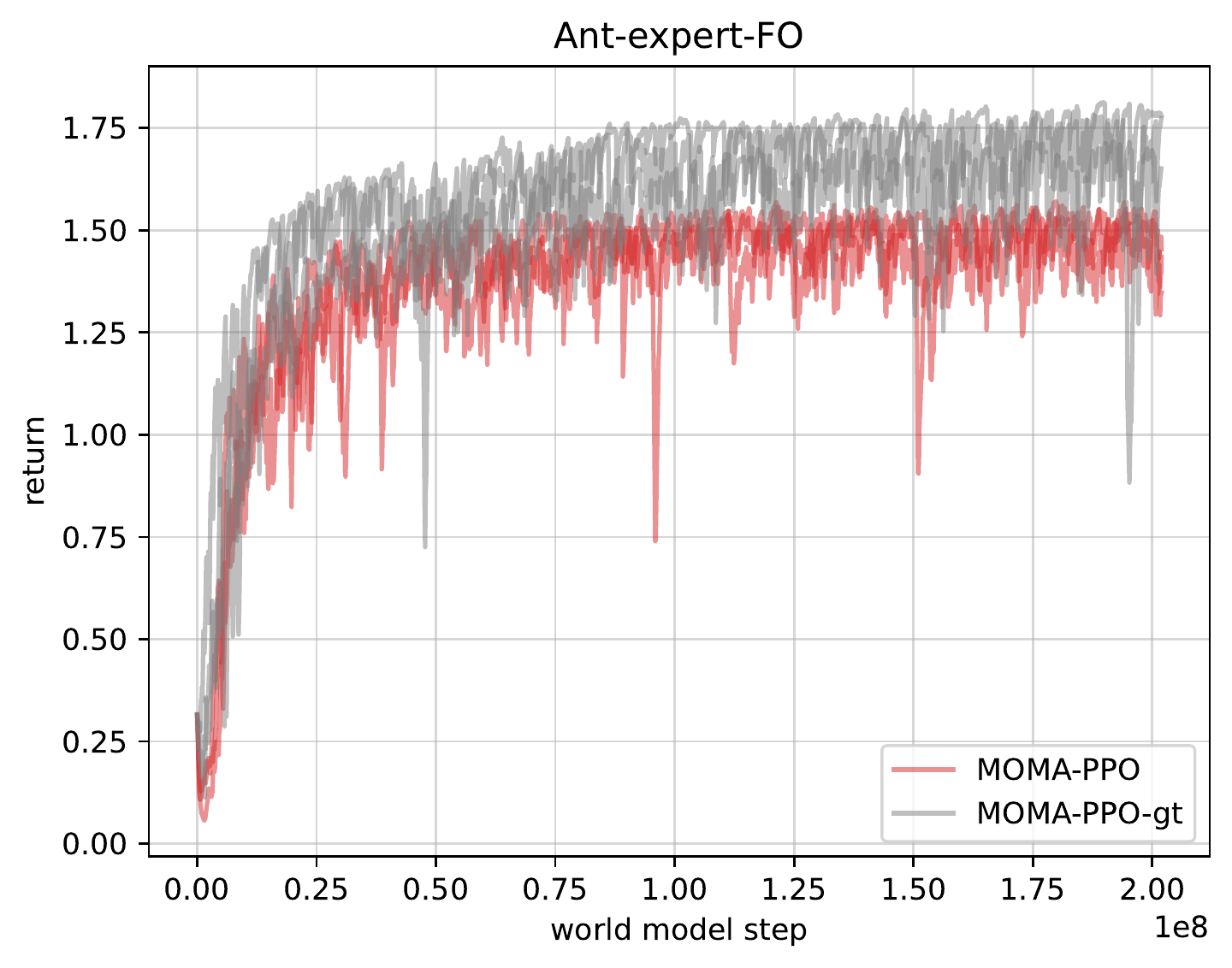} &  \includegraphics[width=0.5\textwidth]{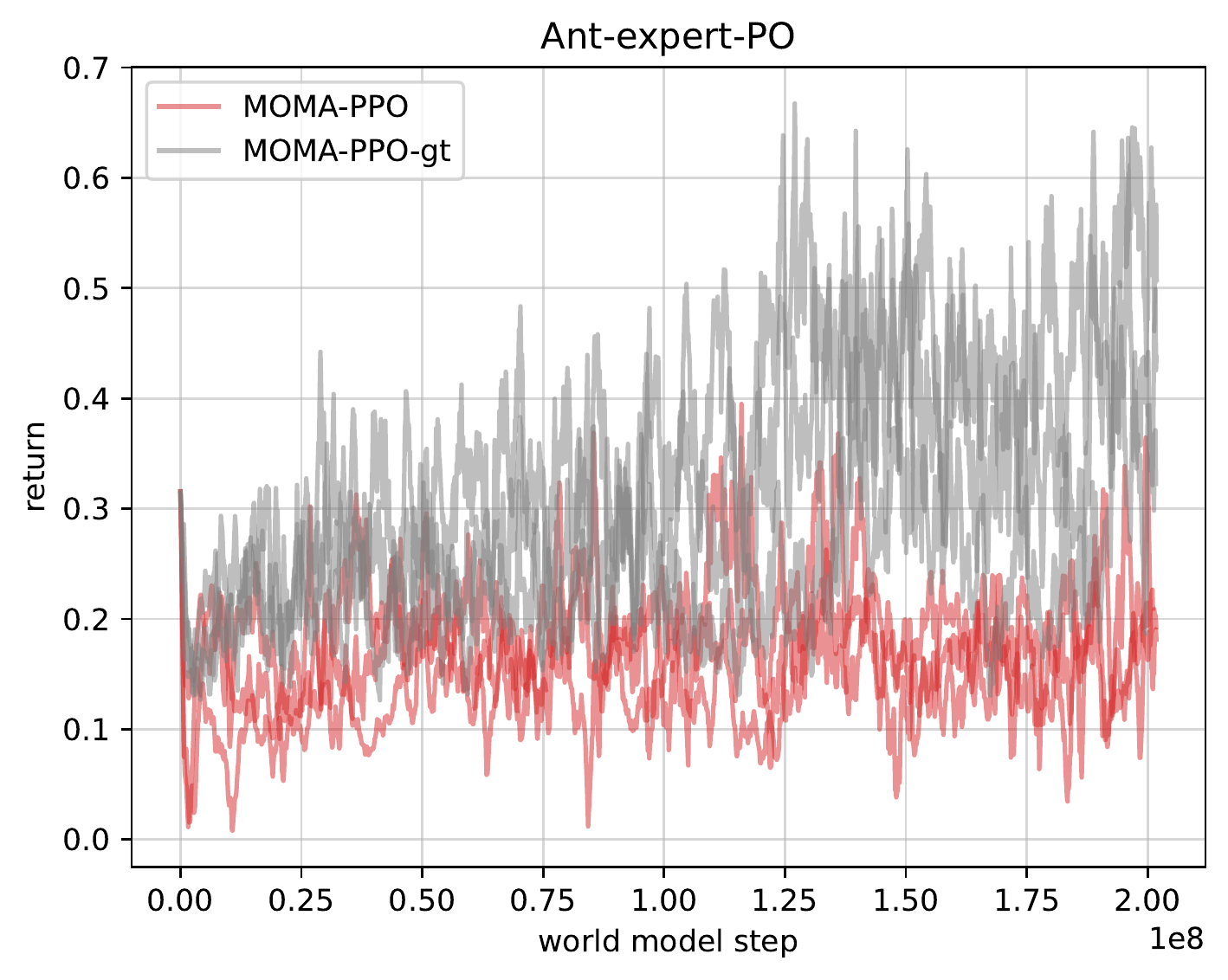}
    \end{tabular}}
    \caption{Impact of using the ground truth world model vs. the learned world model. ``gt" stands for ``ground truth" and means that the corresponding runs use the ground truth world model. The learning curves are with respect to generated transitions (either from the ground truth or the learned world model). Here we show each individual run instead of the usual mean and standard error of the mean.}
    \label{fig:perfect-wm}
\end{figure}

\begin{table}[H]
    \centering
    \caption{Mean scores and standard error of the mean at the end of training with and without the use of the ground truth simulator. Evaluations are over 100 episodes.}
    \begin{tabular}{|c|c|c|c|}
    \hline
   & & \acrshort{moma-ppo} & \acrshort{moma-ppo}-gt \\ 
    \hline
\multirow{4}{*}{(\acrshort{fo})} & ant-random  &  0.52 $\pm$ 0.07  & \textbf{ 1.17 $\pm$ 0.03}  \\ 
& ant-medium  &  1.29 $\pm$ 0.06  &  \textbf{1.68 $\pm$ 0.03}  \\ 
& ant-full-replay  &  1.42 $\pm$ 0.07  &  \textbf{1.66 $\pm$ 0.03}  \\ 
& ant-expert  &  1.49 $\pm$ 0.01  &  \textbf{1.71 $\pm$ 0.04}  \\ 
\hline
\multirow{4}{*}{(\acrshort{po})} & ant-random  &  0.42 $\pm$ 0.05  &  0.47 $\pm$ 0.01  \\ 
& ant-medium &  0.54 $\pm$ 0.19  &  0.81 $\pm$ 0.20  \\ 
& ant-full-replay &  0.46 $\pm$ 0.10  &  \textbf{0.84 $\pm$ 0.07}  \\ 
& ant-expert &  0.18 $\pm$ 0.00  & \textbf{0.43 $\pm$ 0.06}\\
\hline
    \end{tabular}
    \label{tab:perfect-wm}
\end{table}

Figure~\ref{fig:perfect-wm} and Table~\ref{tab:perfect-wm} compare using a learned world model with having access to the ground truth simulator to generate the rollouts. It appears that MOMA-PPO performance can be further improved provided that we learn better world models.

\begin{figure}[H]
    \centering
    \resizebox{1\textwidth}{!}{
    \begin{tabular}{cc}
        \includegraphics[width=0.5\textwidth]{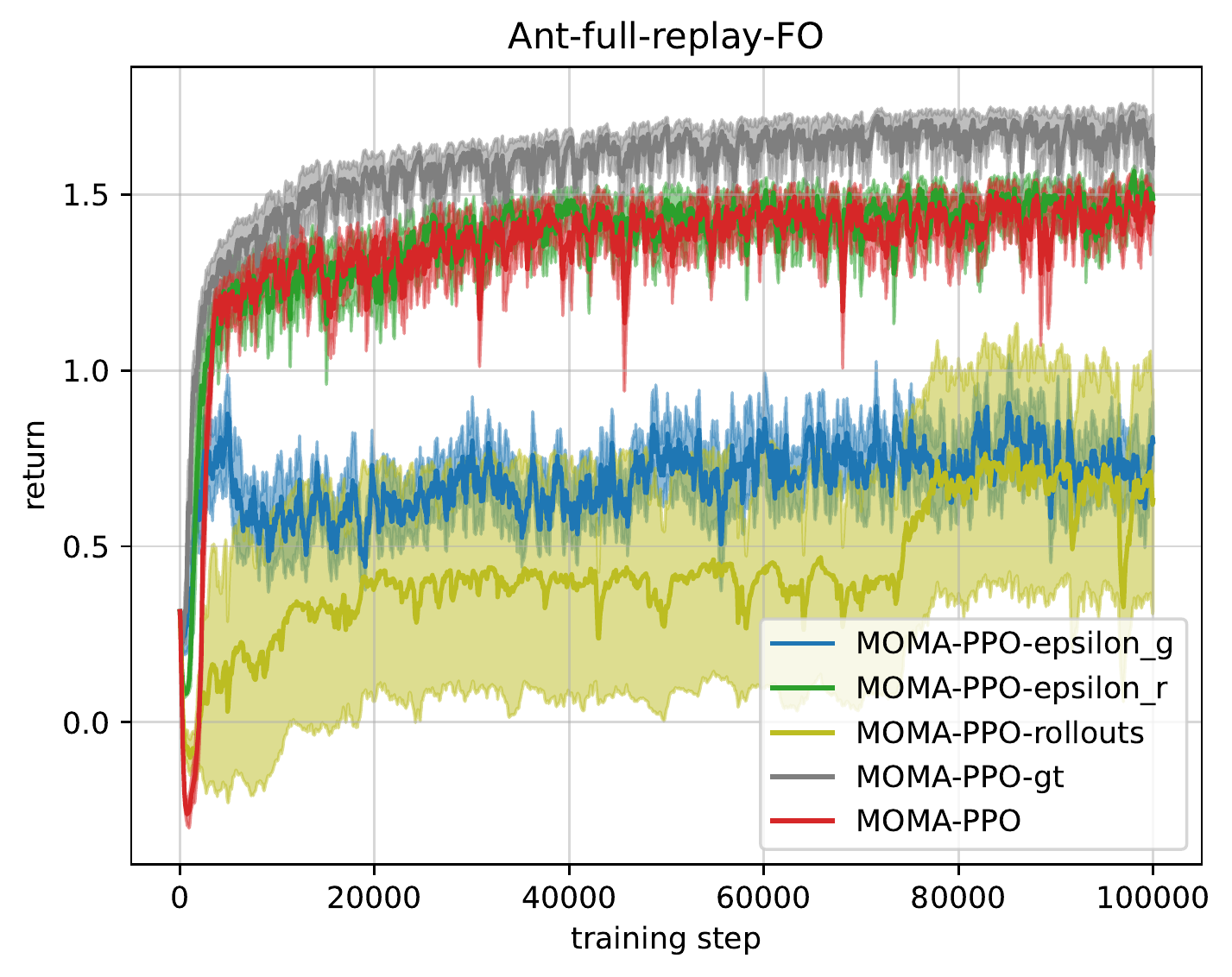} &  \includegraphics[width=0.5\textwidth]{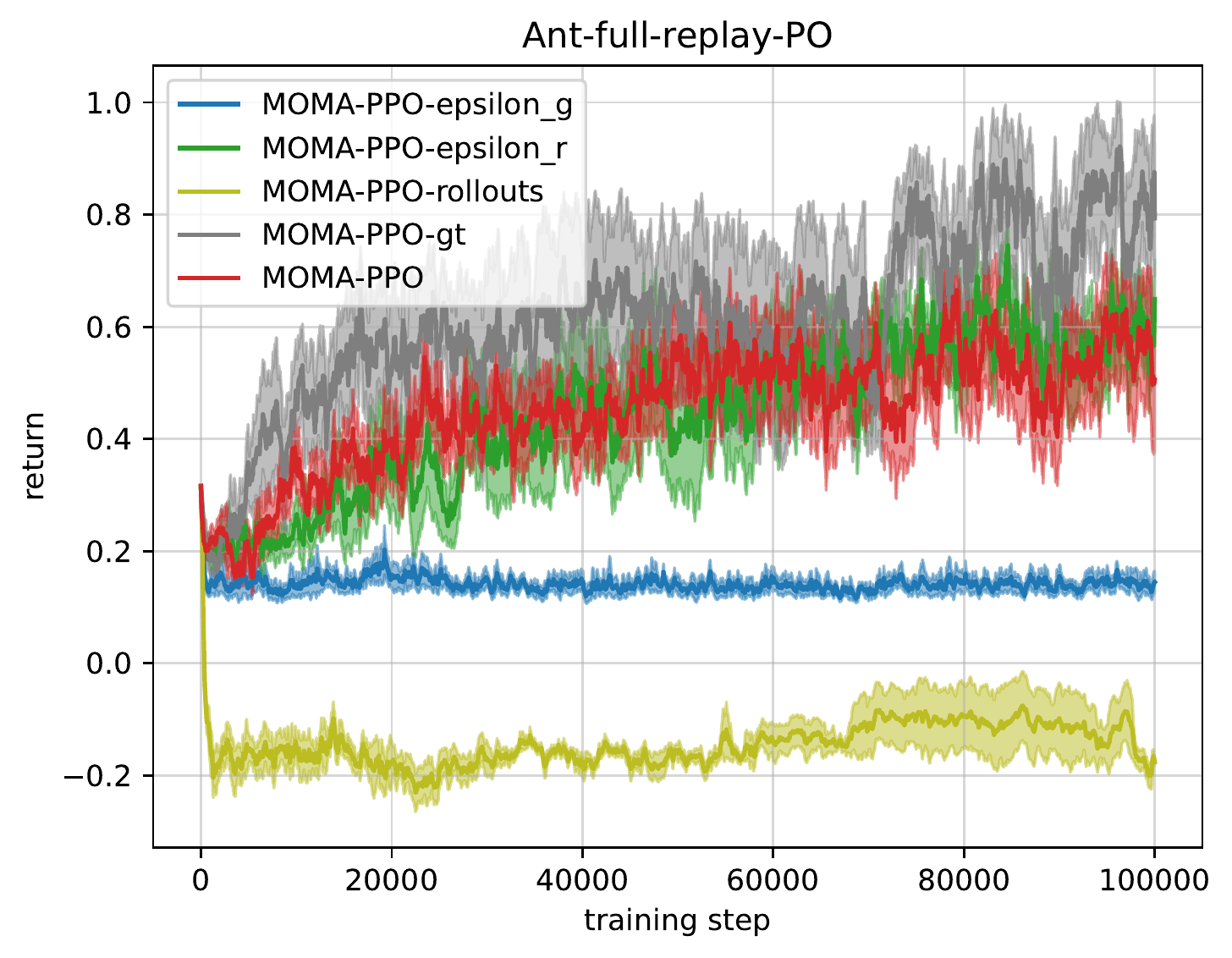}
    \end{tabular}}
    \caption{Impact of the different uncertainty-based techniques to prevent model exploitation by the learning algorithm. ``gt" stands for ``ground truth" and means that the corresponding runs use the ground truth world model. ``epsilon\_g" and ``epsilon\_r" respectively means that we set $\lambda_g=0$ and $\lambda_r=0$. ``rollouts" means that we removed the adaptive rollouts component (i.e. $l_\epsilon = \infty$). Mean and standard error of the mean on three seeds.}
    \label{fig:ablations}
\end{figure}

Figure~\ref{fig:ablations} shows ablations on MOMA-PPO. It appears that $\lambda_r$ penalty can be removed without hurting performance, indeed it is already encapsulated in $\lambda_g$. The other components of MOMA-PPO such as $\lambda_g$ uncertainty penalty on the reward, or the adaptive rollouts that terminate if the uncertainty crosses a threshold, are mandatory to reach satisfactory performance. Note that running experiments without clipping the world model predictions to the datasets' bounding box was impossible as the magnitude of the predicted state exploded after a few rollout steps. This could be mitigated by reducing the exploration of the PPO policies so that the agents stay closer to the dataset distribution where the world model does not predict such extreme states.

\begin{figure}[H]
    \centering
    \resizebox{1\textwidth}{!}{
    \begin{tabular}{cc}
        \includegraphics[width=0.5\textwidth]{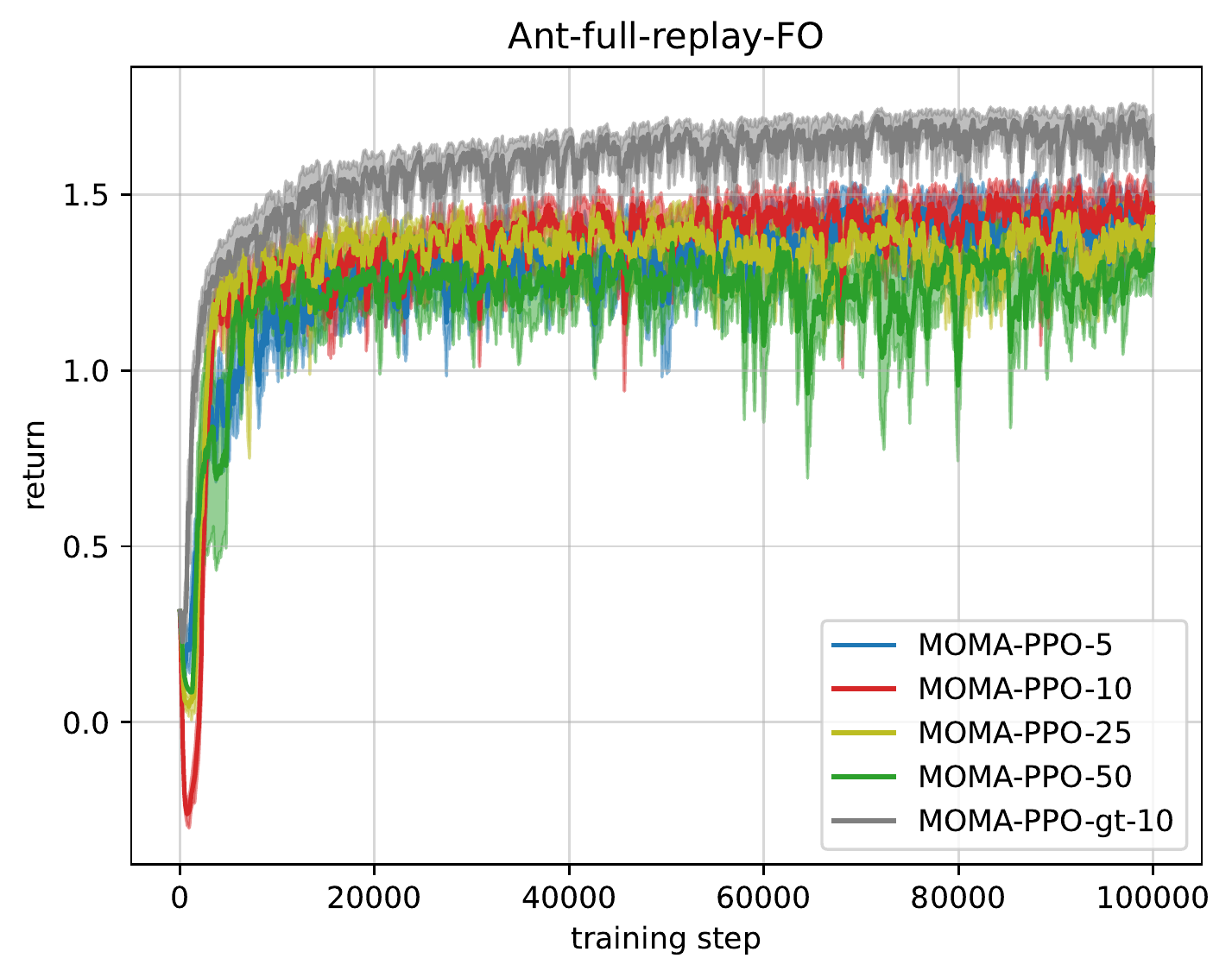} &  \includegraphics[width=0.5\textwidth]{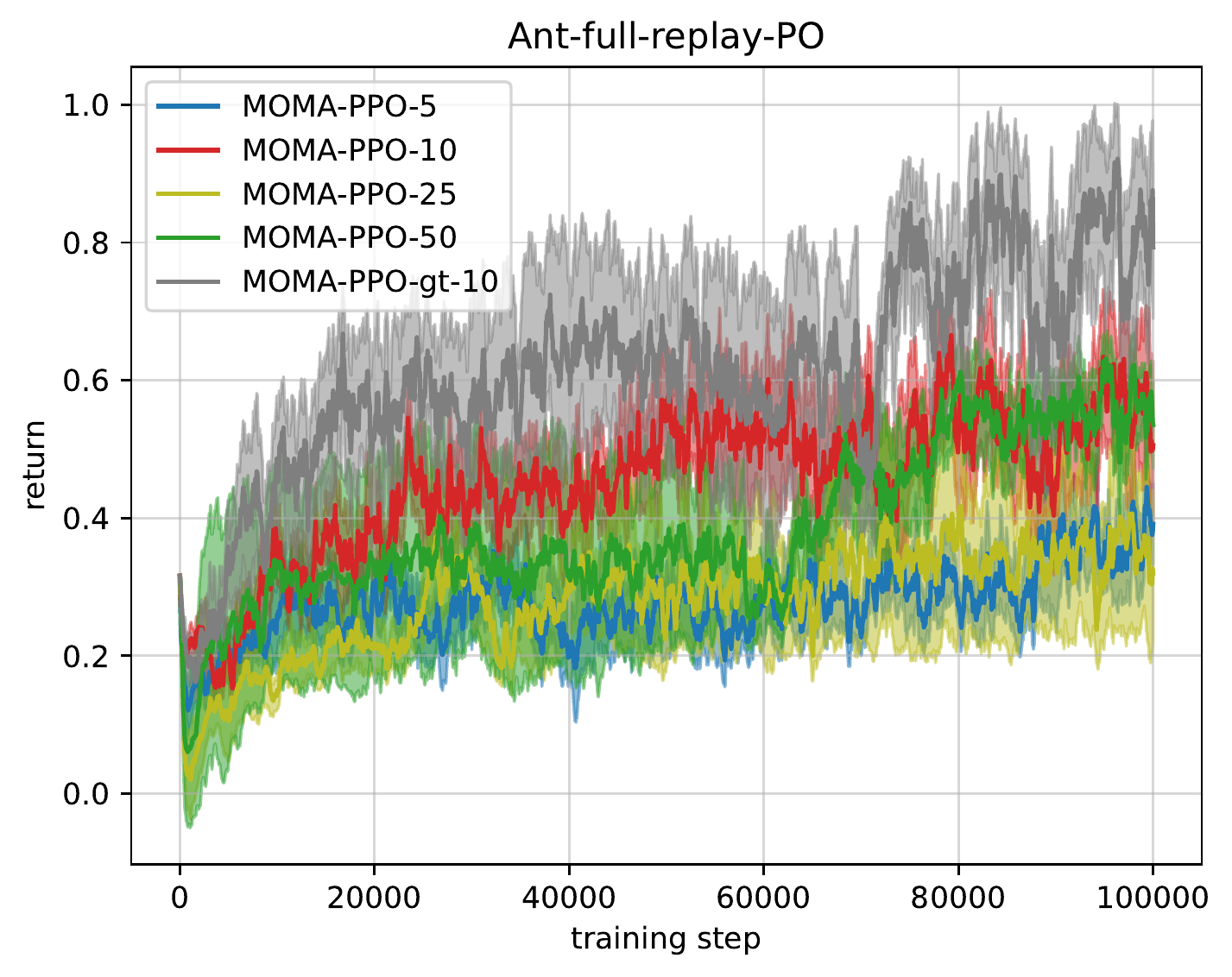}
    \end{tabular}}
    \caption{Impact of different maximum rollout lengths when generating the synthetic interactions. MOMA-PPO-x indicates a maximum rollout length of x. ``gt" stands for ``ground truth" and means that the corresponding runs use the ground truth world model. Mean and standard error of the mean on three seeds.}
    \label{fig:rollout-length}
\end{figure}

Figure~\ref{fig:rollout-length} shows MOMA-PPO trainings for different maximum rollout lengths. It appears that varying the maximum length from 5 to 50 does not significatively impact MOMA-PPO performance on the tasks we investigated. 

\end{document}